\documentclass[a4paper,11pt]{article}
\usepackage{graphicx}

\DeclareGraphicsExtensions{.eps}

\newcounter{tmp_count}

\newenvironment{liste_par_arabic}
{\setcounter{tmp_count}{1} \begin{list}{\textbf{(\arabic{tmp_count})\addtocounter{tmp_count}{1}}}
{\setlength{\itemsep}{\parsep}}}
{\end{list}}

\newenvironment{liste_par_alph}
{\setcounter{tmp_count}{1} \begin{list}{\textbf{(\alph{tmp_count})\addtocounter{tmp_count}{1}}}
{\setlength{\itemsep}{\parsep}}}
{\end{list}}

\title{Mining Heterogeneous Multivariate Time-Series \\ for Learning Meaningful Patterns: \\ Application to Home Health Telecare}

\author{\emph{Florence Duch\^{e}ne$^1$, Catherine Garbay$^1$, and Vincent Rialle$^{1,2}$}\\ \\
$^1$Laboratory TIMC-IMAG, Facult\'{e} de m\'{e}decine de Grenoble, France\\
$^2$Department of Medical Informatics (SIIM), Michallon hospital, Grenoble, France}

\date{}

\begin{document} 

\maketitle

\begin{abstract}
For the last years, time-series mining has become a challenging issue for researchers. An important application lies in most monitoring purposes, which require analyzing large sets of time-series for learning usual patterns. Any deviation from this learned profile is then considered as an unexpected situation. Moreover, complex applications may involve the temporal study of several heterogeneous parameters. In that paper, we propose a method for mining heterogeneous multivariate time-series for learning meaningful patterns. The proposed approach allows for mixed time-series -- containing both pattern and non-pattern data -- such as for imprecise matches, outliers, stretching and global translating of patterns instances in time. We present the early results of our approach in the context of monitoring the health status of a person at home. The purpose is to build a behavioral profile of a person by analyzing the time variations of several quantitative or qualitative parameters recorded through a provision of sensors installed in the home.

\end{abstract}

\section{Introduction}\label{introduction}

In the last years, the increasing amount of stored data with possibly high dimensionality has encouraged researchers to take a great interest in discovering new patterns or building models from large datasets, also referred to as knowledge discovery or data mining. Moreover, many business to scientific applications which serve mainly to support diagnosis and predict future behaviors effectively deal with temporal sequences \cite{roddick}, encouraging the development of the related ``time-series mining'' research field.

In this work we investigate the issue of mining multidimensional and heterogeneous time-series for learning meaningful patterns. This is particularly useful in most monitoring purposes, when dealing with the detection of unusual trends or behaviors of an object or a situation described by the variation of data recorded from several types of sensors or information sources. One application is the monitoring of the health status of a person at home. The aim is to support the caregivers by providing information about unusual trends in the person's behavior observed 
through the variation of quantitative or qualitative parameters monitored at home. 
In that context of detecting bad trends in health status, we aim to learn the person's lifestyle to build a sort of profile, which is sensitive to any critical deviation, and then to detect any unusual behavior in comparison with this profile. This approach toward the decision-making issue is required because it is inconceivable to describe all possible critical situations of any nature and level, just as we do not yet have any way of learning the occurrence of such situations (monitoring of persons getting to critical situations and collecting the corresponding data). 
A learning process is then defined to build a behavioral profile of the person in their activities of daily living, that is to extract and characterize frequent patterns from heterogeneous multivariate time-series recorded in usual conditions of life. The decision-making process must be able to detect unusual behaviors by comparison to this profile. Therefore, the pattern learning process should allow for heterogeneous components defining time-series, as well as for imprecise matches, outliers, stretching, and global translating in time of the sequences corresponding to a same pattern. 

Considering our context also justify the choice of extracting multidimensional patterns related to the person's behavior rather than analyzing individually each parameter monitored at home to make a joint decision about their condition of life. Indeed, the observable parameters are selected as a compromise between: (1) being easily observable and non invasive, and (2) gaining a full appreciation of the person's condition, sensitive to any change in the health status. Therefore, all parameters are closely related one to each other, and their joint variations need to be preserved in multidimensional patterns representative of any usual behavior.
Our objective is then to build a system performing an unsupervised extraction of this kind of temporal patterns within time-series representative of a person's usual conditions of life. Our contribution lies in extending an algorithm for pattern extraction to both multidimensional and heterogeneous time-series, accounting for the large amount of noise possibly present in patterns' instances. We then also need to introduce a similarity measure appropriate to the comparison of multidimensional and heterogeneous time-series.

The rest of the paper is organized as follows. In section \ref{related_works} we present works related to time-series mining, and in section \ref{method_patterns_extraction} our methodology for extracting frequent patterns from heterogeneous multivariate time-series. Then, section \ref{similarity} defines an appropriate similarity measure, and section \ref{patterns_extraction} details the different steps required for pattern discovery and clustering. Section \ref{experimental_results} discusses the early experimental results related to the proposed approach in the context of home health telecare. Finally, section \ref{conclusion} concludes the paper.

\section{Related work}\label{related_works}

Most of the current works dealing with home health telecare are focused either on implementing a generic architecture for the integrated medical information system, on improving the daily life of patients using various automatic devices, specific equipment, and basic alarms, or on providing health care services to patients suffering from specific diseases like asthma, diabetes, cardiac, pulmonary, or Alzheimer's. Rialle \textit{et al.} have presented in \cite{rialle_bb} an overview of projects related to home health telecare. Basic alarms are raised by "smart" sensors or low layers of a local intelligence unit when a problem occurs at a short temporal scale: either one parameter overpasses a critical value (nocturia, pollakisuria, fall, hypertensive crisis, etc.), or a critical scenario involving the value of possibly more than one parameter is recognized (asthma crisis, etc.). Our focus is on the broadcasting of high level alarms about the person's health status, which concern a larger temporal scale. That issue is solved by first learning the daily living habits of a person to be able to detect later unusual situations. That behavioral profile is built by mining heterogeneous multivariate temporal data collected from sensors installed at home for learning meaningful patterns.

According to Antunes \textit{et al.} \cite{antunes}, temporal sequences are related to series of nominal symbols from a particular alphabet, whereas time-series concern continuous, real-valued elements. In this work we are interested in heterogeneous multivariate sequences of time-varying data, referred to as heterogeneous multivariate either time-series or temporal sequences. Time-series mining is an active field of research (see \cite{antunes, roddick} for overviews of temporal data mining). Discovery algorithms for time-series aim at extracting important patterns such as similarities, trends, or periodicity, in a purpose of description or prediction \cite{nanopoulos}. Pattern discovery in time-series is useful for temporal sequences synthesis \cite{hong_b}, as well as for learning tasks like association rules mining \cite{das_b, hoppner}, classification \cite{keogh}, unsupervised clustering \cite{vlachos}.

By analogy with non sequential domains and because of the exponentially large set of possible subsequences considering temporal sequences, time-series mining used to serve a learning task is sometimes referred to as ``feature mining'' \cite{kudenko, lesh}. Considering non-sequential domains, feature selection corresponds to finding an optimal space of size $m$ from the full $d$-dimensional feature space, where ideally $m\ll d$. In sequential domains, ``feature selection aims to select the best subset of sequential features'' \cite{lesh}, that is, the most relevant subsequences regarding the decision purpose. Time-series mining then acts as a preprocessor to construct the best subset of sequential features used to feed into learning algorithms \cite{lesh_b}. This is particularly useful to improve learning performances when time-series contain both pattern and non-pattern signals, like in \cite{hong_b}. According to \cite{lesh}, the selection criteria for feature mining include that features should be frequent, distinctive of at least one class, and that feature sets should not contain redundant features, that is subsequences.

Pattern discovery in time-series may be either (1) supervised -- that is finding patterns described by empirical knowledge or similar to a given ``query sequence'' \cite{agrawal, keogh, lee} -- or (2) unsupervised -- that is finding recurrent patterns without any prior knowledge about the regularity of the data under study \cite{chiu, hong, hong_b}. Lin \textit{et al.} \cite{lin} have introduced the notion of ``time-series motifs'' considering the unsupervised issue of finding previously unknown, frequently occurring patterns in time-series. These specific patterns are also referred to as ``primitive shapes'' \cite{das_b} or ``frequent temporal patterns'' \cite{hoppner}.

The techniques used for time-series mining vary according to the application, regarding the characteristics of both the temporal sequences under study and the expected patterns: degree of variability in the values, allowed transformations between instances of a same pattern, possible stretching in time. For instance, Hong \textit{et al.} have experimented training recurrent neural networks for an unsupervised extraction of multi-temporal sequence patterns \cite{hong}. However, this method suffers from noisy data. The use of finite state machines \cite{hong_b} may give out good results, which may however dramatically decrease as the dimensionality increases - that is the number of states. Chiu \textit{et al.} \cite{chiu} have implemented in the context of times-series motifs extraction an efficient algorithm based on random projections initially proposed by Buhler and Tompa to find motifs in nucleotide sequences \cite{buhler}. Although they only deal with one-dimensional time-series and then do not address the issue of heterogeneous multivariate time-series, this \emph{projection} algorithm is actually interesting because of the rapid extraction of approximate results and the efficiency even in the presence of noise or ``don't care'' symbols. However this method, as implemented in \cite{chiu}, does not allow for stretching in time between motifs instances.

Our objective is then to extend feature mining and learning from time-series to the unsupervised extraction of heterogeneous multivariate time-series motifs for learning a behavioral profile. We extend the use of the \emph{projection} algorithm for feature mining in our noisy, heterogeneous and multi-dimensional context. The aim is to extract the most relevant features -- that is, subsequences in time-series domain -- to feed into a clustering algorithm for motifs identification.
As an application, we focus on profiling the daily living habits of a person from data recorded using a provision of sensors installed in their home.

\section{A methodology for mining heterogeneous multivariate ti\-me-series}\label{method_patterns_extraction}

\subsection{Problem entity\\\emph{Pattern extraction in home health telecare}}\label{problem_entity}

Solving any complex decision making system requires to well specify the purpose and context of the decision. Dealing with several levels of details -- like different levels of knowledge, data accuracy, decision -- particularly requires to carefully set the needs, requirements and constraints of the system, so that the decision making matches the defined purpose at most.
Setting up a decision making system -- as motifs extraction -- should then be considered as part of a problem solving scheme \cite{duchene_b}, including (see figure \ref{fig_pb_solving}):
\begin{figure}[t]
\centering
\includegraphics[width=3.5in]{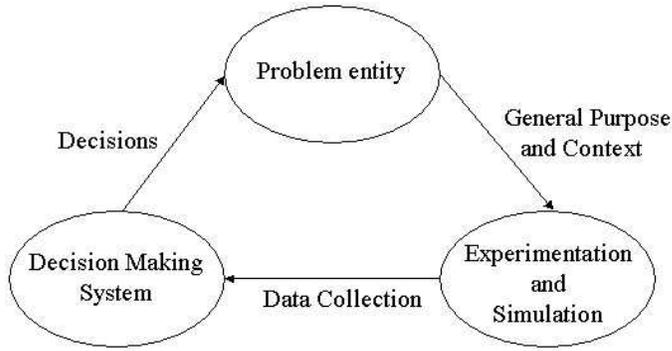}
\caption{\small \textbf{The recursive steps of a problem solving scheme.}}\label{fig_pb_solving}
\end{figure}
\begin{enumerate}
        \item \textbf{Defining the context and the general purpose of the decision issue.} This aims at narrowing and specifying the space of information and knowledge to consider by answering questions like: what are the relevant observations to set up ? or which level of detail to consider? or what are the performance expected for the problem solving ?
        \item \textbf{Collecting or generating data related to that context.} This data collection is led by contextual information related to the general purpose and context of the decision issue. Collecting large sets of representative data may be quite challenging in some applications. Setting up a simulation process \cite{duchene_b} is then very useful as a first step of setting up a decision system to prevent from the lack of experimental data. Simulation also allows to get a full view of data potentially recorded in the context of study, by varying the parameters of the simulation process, so that the performance of decision making systems are better evaluated.
        \item \textbf{Testing appropriate methods of decision making to solve the problem.} Data collected from experiments or generated by a simulation process are used as inputs of the decision-making system. The sensitivity and specificity related to these algorithms must match the problem requirements.
\end{enumerate}
Once a decision process has been implemented and experimented, the results of matching between the outputs of the decision making process and the problem requirements may entail to refine one or more steps in order to get better sensitivity and specificity. For instance, more precision in the values of the observed parameters may be required in case of low sensitivity, or conversely, in case of low specificity.
The problem solving scheme must also integrate validation at each stage, like face validation, with experts, or mathematical and statistical validation. 

Defining the problem entity -- that is the context and purpose of the decision -- have some consequences on both setting up the experimental context and building the decision making system. Considering the decision making purpose and its complexity, we can specify some key parameters involved at these stages of any problem solving scheme, as follows:
\begin{itemize}
        \item \textbf{Level of knowledge available.} This aims at identifying all knowledge possibly available and relevant to decision making, that is (a) \textit{a priori} knowledge -- including intuitive and academic knowledge -- and (b) knowledge extracted from experimental data sets. Fusion of several kinds of knowledge is commonly used to deal with complexity and heterogeneity. The lack of knowledge related to a specific issue may also require to rely on various informational sources.
        \item \textbf{Level of details required.} Specifying the level of details required to deal with an issue is crucial to select: (a) appropriate knowledge, and consequently the level of experimental data collection and representation, and (b) relevant algorithms to solve the problem. There is a compromise to be found between the necessity to save the complexity of phenomena, and the restriction to a level of detail relevant to the problem, that is meeting the decision's purpose.
        \item \textbf{Level of performance expected.} The performance is defined using parameters like the sensitivity and specificity of the decision making system, such as an acceptable time to decision.
\end{itemize}
In our context of home health telecare, the issue of extracting patterns representative of a person's daily behavior at home is a \emph{high level issue}. The ultimate goal is not to interpret precisely a problem that occurred at home, but to set up the context of occurrence of any change in the behavior. Therefore, the pattern extraction aims at identifying recurrent behaviors occurring at the scale of long time intervals, from about thirty minutes to several hours. The ``right levels'' to deal with that issue are detailed in the following paragraphs:
\begin{itemize}
        \item \textbf{Level of knowledge.} Learning about daily behavioral profiles must be performed on individual basis, since behavioral profiles are specific of a person's physiological status and habits. Consequently, there is only a few \textit{a priori} knowledge related to our decision issue. The decision-making system is then based on a set of data recorded at home and in real-time from a provision of sensors. In order to gain a full appreciation of the person's condition, sensitive to any change in the health status, data may be collected from different classes of sensors: (1) activity (location, posture, etc.), (2) environment (temperature, use of doors, window, lighting, etc.), and (3) physiology (heart rate, blood pressures, weight, etc.). The selection of relevant parameters is also constrained by many ethical, social, and individual issues: respect of privacy, confidentiality of data, ease of use and unobtrusiveness of input devices installed in the home.
        \item  \textbf{Level of details.} Dealing with a high level issue, the decision making system may not require a high degree of detail and accuracy as regards the data involved in the process, that is data collected from sensors in real-time. Moreover, the experimental records may require a high level of representation to highlight their global trends, removing minor variations that are insignificant at our observation scale.
        \item \textbf{Level of performance.} The decision system needs to fit the general purpose of monitoring: the detection of all usual patterns (sensitivity) combined with a low rate of false alarms (specificity) -- that is identifying patterns that do not correspond to usual behaviors, with an acceptable time to detection.
\end{itemize}

\subsection{Experimental context\\\emph{Some guidelines for pattern extraction}}\label{guidelines}

An appropriate experimental context is set according to the purpose and requirements of the problem. The characteristics of the data produced in that context induce some guidelines for pattern extraction. Looking for meaningful patterns representative of human behaviors -- the activities of daily living of a person at home -- from heterogeneous data collected from a provision of sensors, the decision making-system must be able to address the following issues:
\begin{itemize}
        \item \textbf{Multivariate time-series.} Relevance for dealing with time-varying objects or situations described by several parameters.
        \item \textbf{Heterogeneous components.} Capacity of handling qualitative as well as quantitative parameters in a coherent way to describe an object or a situation.
        \item \textbf{Mixed time-series.} Ability to learn from sequences containing both pattern and non-pattern data. Human behaviors captured in daily life indeed contain highly casual as well as regular motions.
        \item \textbf{Imprecise matches.} Capacity to discover ``high-level patterns'', that is to focus on the global trends embedded in the data despite the strong presence of noise between the instances of expected patterns, especially when considering human behaviors.
        \item \textbf{Outliers.} Capacity to preserve the detection accuracy despite the presence of outliers in subsequences corresponding to frequent patterns. That may be due to anomaly in the sensor or attributed to human failure or disruption.
        \item \textbf{Translation in time.} Ability to detect patterns translated in time: similar behaviors may occur at any time.
        \item \textbf{Stretching in time.} Ability to detect patterns of different lengths: dealing with human behaviors, a same activity does not always last the same duration.
\end{itemize}

\subsection{Decision-making system\\\emph{Methodology of pattern extraction}}\label{methodology}

The decision-making system aims at extracting meaningful temporal patterns from heterogeneous multivariate time-series. The patterns should correspond to usual behaviors of a person at home. Given that activities of daily living are specific to a given subject, we need a completely unsupervised learning approach, that may not be driven by prior knowledge about the patterns corresponding to living habits. Unsupervised time-series mining is usually made up of several consecutive steps to perform coarse-to-fine feature extraction \cite{chiu, hong_b, hong, lee}. The principle is first to roughly restrict the feature space combining techniques like time-series representation, random projections \cite{chiu}, state-based temporal signal modelling \cite{hong_b}, recurrent neural nets training \cite{hong}, and then to identify more precisely the relevant subsequences for the learning task based for instance on specific constraints or similarity thresholds on the subsequences. As a consequence, this implies to define relevant similarity measures for the given time-series. 
Antunes \textit{et al.} \cite{antunes} define temporal data mining as a process including three main steps:
\begin{itemize}
        \item \textbf{Representation of temporal sequences.} Preprocessing, representation, and modelling of the data sequences that need to be applied before actual data mining operations take place (transformation, discretization, generative models building).
        \item \textbf{Similarity measure for sequences.} Definition of an appropriate similarity measure according to the characteristics of the time-series.
        \item \textbf{Mining operations.} Application of models and representations to the actual mining problem (association rules mining, classification, unsupervised clustering, prediction).
\end{itemize}

\begin{figure}[t]
\begin{center}
\includegraphics[width=\textwidth]{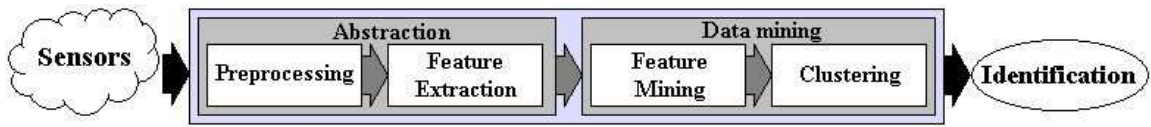}
\end{center}
\caption{\small \textbf{Pattern recognition system designed in the context of mining \emph{low-level} time-series for identifying \emph{high-level} patterns.}}\label{fig_patterns_extract}
\end{figure}

\noindent Our approach differs from the one of Antunes \textit{et al.} in the way we specify each of these three steps. Considering a complex issue involving a large scale from the level of details embedded in raw data to the decision level, we need to refine the definition of representing and mining temporal sequences: (a) Representation needs then to be defined as a step of \emph{abstraction}, to get from raw data a level of information that better deals with the decision purpose -- preprocessing and feature extraction ; and (b) mining operations are divided into two consecutive steps -- feature mining and clustering -- to progressively focus on the most appropriate features to pattern extraction. We then propose a new general design of pattern recognition systems in the case of dealing with large temporal data sets, as shown on figure \ref{fig_patterns_extract}. The following paragraphs refine in that context the three steps defined by Antunes \textit{et al.}.

\subsubsection*{Representation}
Considering the extraction of high level patterns from time-series (see \S \ref{problem_entity}), the step of \textbf{representation} is not a simple preprocessing of data. Once represented, time-series must fit the level of details required by the decision system. We then define the representation as a step of \emph{abstraction} of raw data to capture a higher level of information most appropriate to the decision's purpose.
The aim is to get a synthetic representation of the sequential data, meaningful in terms of identifying the activities of daily living of a person at home.
Given than an activity can be described as a succession of elementary \emph{``actions''}, each of them being performed for a certain duration, we aim at representing raw time-series by sequences of meaningful symbols, each symbol representing the person carrying out a given ``action'' for a certain time. That implies to gather along time successive data records corresponding \textit{a priori} to a same ``action'', that is whose sequence present no significant temporal variations. Time-series are then abstracted as sequences of symbols -- some multidimensional vectors, each of them synthesizing a ``stationary'' state of the monitored parameters during a certain period of time.

\subsubsection*{Mining operations}
Dealing with mixed time-series, embedding both pattern and non-pattern data subsequences (see \S \ref{guidelines}), we propose to divide the definition of \textbf{mining operations} into two consecutive steps to progressively better match the decision level:
\begin{enumerate}
        \item \textbf{Feature mining.} Selection of the most meaningful features (that is, subsequences), so called \emph{tentative motifs}, considering the purpose of frequent patterns extraction and classification. These subsequences are used as features to feed into a classification algorithm.
        \item \textbf{Clustering.} Unsupervised classification of the \emph{tentative motifs} into meaningful clas\-ses whose representative sequences are called \emph{time-series motifs}. 
\end{enumerate}
In our experimental context, tentative motifs should be representative of a person's repetitive behaviors, and a \emph{motif} is then defined as a meaningful class of tentative motifs, representative of any typical activity of the person.
Because we need a completely unsupervised learning approach, both \emph{feature mining} and \emph{clustering} must be unsupervised. 

\subsubsection*{Similarity measure}
Considering our level of complexity, a similarity measure is required at the two stages of representing and mining time-series:
\begin{enumerate}
        \item \textbf{Representation.} The purpose of representation is to synthesize in a single symbol subsequences whose successive vectors share similar values, possibly for different durations, and therefore representative of a same ``action'' performed along the corresponding time. We then need to roughly evaluate the proximity of successive vectors and their relevance to be abstracted in one symbol describing a continuous same type of action performed. For the sake of robustness and efficiency, a discretization step is first of all applied to quantitative parameters. Given the fact that we are looking for global trends, a low approximation of the actual distance between a subsequence of vectors and its corresponding ``mean vector'' is sufficient. In case these subsequences are roughly similar, we can then assume the ``mean vector'' well represent the subsequence for the corresponding duration. As a consequence, we propose to use a \emph{discrete minimum distance} for that purpose (see \S \ref{mindist}). 
        \item \textbf{Mining operations.} Once the possible locations of patterns have been identified within the original time-series, a similarity measure between these subsequences is required for (a) \emph{Feature mining}, to decide whether or not they are effectively similar enough to be considered as a possible pattern, and (b) \emph{Clustering}, to classify all these extracted subsequences into meaningful groups in terms of characterizing the activities of daily living. At this stage we need to compute an \emph{actual distance} relevant to heterogeneous multivariate time-series (see \S \ref{seqdist}).
\end{enumerate}
In that context, specifying a similarity measure then includes defining the following elements: (1) homogeneous distance for heterogeneous components, (2) \emph{actual distance} between heterogeneous multivariate time-series, and (3) \emph{minimum distance} between time-series.

The next sections details, first, the definition of required similarity measures and, second, the proposed approach for identifying time-series motifs, including (1) representation of time-series, (2) \emph{feature mining}, for tentative motifs discovery, and (3) \emph{clustering}.

\section{Similarity measure}\label{similarity}

\begin{figure}[t]
\begin{center}
\includegraphics[height=7cm]{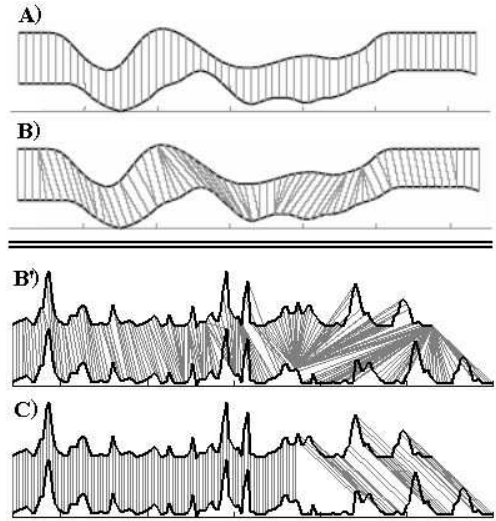}
\end{center}
\caption{\small \textbf{Pairs of points considered as similar and associated when computing the distance between time-series.}}\label{fig_similarity_mes}
\small The graphs represent the results of associations when computing (A) Euclidian distance, (B \& B') Dynamic Time Warping (DTW) distance, and (C) distance based on the longest common subsequence (LCSS).
\begin{itemize}
\item The comparison of points associations between A) and B) shows the better efficiency of DTW distance over an Euclidian distance to deal with possible distorsions in the time axis.
\item The comparison of B') and C) highlights the better efficiency of LCSS over DTW distances to support the presence of outliers.
\end{itemize}
\end{figure}

Various similarity models have been successfully used to compare temporal sequences, as illustrated on figure \ref{fig_similarity_mes} and detailed below. The simplest approach typically used to define a similarity function is based on the Euclidian distance, or some extensions to support various transformations such as scaling or shifting. Chui \emph{et al.} \cite{chiu} have used it successfully for extracting one-dimensional time-series motifs in some specific cases. However, this model cannot deal with outliers and is very sensitive to small distorsions in the time axis (see figure \ref{fig_similarity_mes}, case \textbf{A)}).
Another approach is to use the \emph{Dynamic Time Warping (DTW)} distance which allows stretching in time and comparing time-series of different lengths \cite{keogh_dtw,kruskall_dtw} (see figure \ref{fig_similarity_mes}, case \textbf{B)}). However, a great amount of outliers still results in very large distances, even though the difference may be found in only a few points (see figure \ref{fig_similarity_mes}, case \textbf{B')}).
Non-metric techniques have then been introduced and efficiently used to better deal with noisy data \cite{agrawal,das,vlachos}. The idea is to capture the intuitive notion that ``two sequences should be considered similar if they have enough non-overlapping time-ordered pairs of subsequences that are similar'' \cite{agrawal}. This refers to finding the \emph{Longest Common Subsequence (LCSS)} between two time-series. This approach allows for outliers, different scaling factors, and baselines (see figure \ref{fig_similarity_mes}, case \textbf{C)}).

However, the above works mainly deal with low dimensional (from one to three dimensional) time-series and do not address the issue of heterogeneous components (quantitative or qualitative) describing a moving object. Considering heterogeneous multivariate time-series in a particularly noisy context, our objective is then to extend the \emph{LCSS} approach to heterogeneous multivariate time-series. We first need to define a coherent distance between points whatever the type of parameter. Then we detail its integration in computing the \emph{actual distance} between heterogeneous multivariate time-series, which is used for \emph{mining operations}. At last, we use these definitions to extend the approach proposed in \cite{chiu} for computing a \emph{minimum distance} between time-series, which is used for time-series \emph{representation}.

\subsection{Homogeneous distance between heterogeneous components}\label{ptsdist}

We would like to allow the description of an object using several parameters of the following possible types:
\begin{itemize}
        \item Quantitative
        \item Ordered qualitative
        \item Unordered qualitative 
\end{itemize}
The simplest way of insuring the coherence of the similarity measure is to make the distances between two values range from 0 to 1 for each type of parameters. Let $a$ and $b$ be two values of a given parameter, and $d(a,b)$ the distance between these two values. In case of a qualitative parameter, let $v$ be the number of variates, the possible values being then the integers from 1 to $v$. According to the parameter's type, $d(a,b)$ is defined as follows:
\begin{equation}\label{dist_quant}
d(a,b) = \left|a - b\right|,
\end{equation}
\begin{equation}\label{dist_qual_ord}
d(a,b) = \frac{\left|a - b\right|}{v-1},
\end{equation}
\begin{equation}\label{dist_qual_unord}
d(a,b) = min(\left|a - b\right|,1).
\end{equation}

The equations (\ref{dist_qual_ord}) and (\ref{dist_qual_unord}) are used respectively for ordered and unordered qualitative parameters. In the case (\ref{dist_quant}) of quantitative parameters, getting a distance between 0 and 1 requires a step of normalization so that the possible values range from 0 to 1. We use a min-max normalization, where the minimum and maximum bounds are defined from experts or using statistical analysis of training sets. All values are then restricted to these bounds, lower and upper values being interpreted as noisy or erroneous. Let $X_{min}$ and $X_{max}$ be respectively the minimum and maximum bounds for the values $x$ of a given parameter $X$. We define the normalized value $norm(x)$ of $x$ as follows:
        \[norm(x) = \frac{max\left(0,min\left(x,X_{max}\right)-X_{min}\right)}{X_{max}-X_{min}}\]

\subsection{Actual distance between time-series}\label{seqdist}

\begin{figure}[h]
\begin{center}
\includegraphics[width=3.5in]{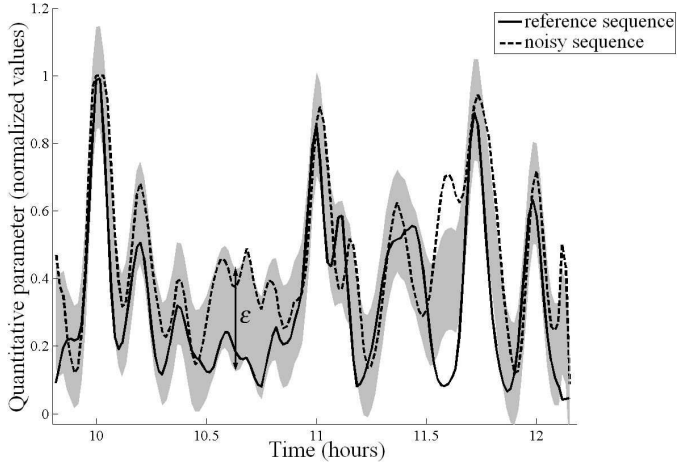}
\end{center}
\caption{\small \textbf{The notion of the \emph{LCSS} matching within a region of $\epsilon$.}}\label{fig_epsilon}
\small Comparing the trajectories point to point along the time axis, the pairs both within the gray region can be matched.
\end{figure}

\begin{figure}[h]
\begin{center}
\includegraphics[width=3.5in]{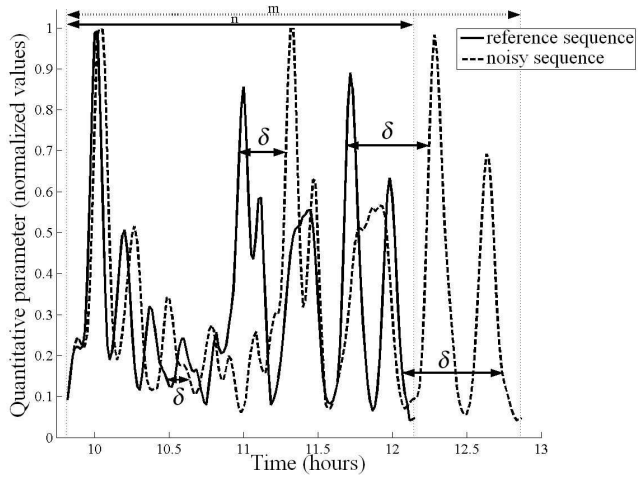}
\end{center}
\caption{\small \textbf{The notion of the \emph{LCSS} matching within a region of $\delta$.}}\label{fig_delta}
\small The points of two trajectories can be matched if the time interval is under the maximum authorized value for $\delta$.
\end{figure}

The similarity function between trajectories is based on the \emph{Longest Common Subsequence (LCSS)}, already used by Vlachos \emph{et al.} \cite{vlachos} in the context of multidimensional (generally two or three dimensional) time-series of quantitative data. Indeed, dealing with noisy data (see \ref{guidelines}) have proved to be better handled using non-metric \cite{agrawal, das, vlachos}, based on the \textit{LCSS}, than metric distances, like the Euclidean distance \cite{chiu}, or the \emph{Dynamic Time Warping (DTW)} \cite{keogh_dtw, kruskall_dtw}. Using \emph{LCSS}, the overall idea is to count the number of couple of points from two sequences $A$ and $B$ that matches according to a predefined matching threshold $\epsilon$, and when going through the temporal sequences (see figure \ \ref{fig_epsilon}). One point can never be associated twice to a point of the other sequence, so that the maximum number of associations is the minimum length of the two sequences. Another constant $\delta$ controls how far in time we can go in order to match points from one trajectory to the other one (see figure\ \ref{fig_delta}).

We assume objects are points moving in a $p$-dimensional space $(x_1,\ldots,x_p)$.
Let $A$ and $B$ be the two trajectories of moving objects with size $n$ and $m$ respectively:
        \[A = ((a_{x_1,1},\ldots,a_{x_p,1}),\ldots,(a_{x_1,n},\ldots,a_{x_p,n})),\]
        \[B = ((b_{x_1,1},\ldots,b_{x_p,1}),\ldots,(b_{x_1,m},\ldots,b_{x_p,m})).\]
For a trajectory $A$, let $Head(A)$ be the sequence:
        \[Head(A) = ((a_{x_1,1},\ldots,a_{x_p,1}),\ldots,(a_{x_1,n-1},\ldots,a_{x_p,n-1})).
\]
Given an integer $\delta$ and a real number $0<\epsilon<1$, the similarity function $LCSS_{\delta,\epsilon}(A,B)$ is defined using the recurrent algorithm (\ref{constraints_LCSS}) \cite{vlachos}. $N$ and $M$ are the size of the sequences $A$ and $B$ respectively at the first step of the recurrent algorithm. 
\begin{equation}\label{constraints_LCSS}
        LCSS_{\delta,\epsilon}(A,B) = \left\{
        \begin{tabular}{ll}
        0 & \mbox{if $A$ or $B$ is empty},\\ \\
        \multicolumn{2}{l}{$1+LCSS_{\delta,\epsilon}(Head(A),Head(B))$,}\\
        & \mbox{if $d\left(a_{x_k,n},b_{x_k,m}\right)<\epsilon$,\ $\forall1\leq k\leq p$},\\
        & \mbox{and $\left|n-m\right|\leq\delta \mbox{\ and\ } \left|N-n-M+m\right|\leq\delta$},\\ \\
        \multicolumn{2}{l}{$max\left(LCSS_{\delta,\epsilon}(Head(A),B),LCSS_{\delta,\epsilon}(A,Head(B))\right)$}\\
        &\mbox{otherwise}.
        \end{tabular}
        \right.
\end{equation}
Our similarity measure differs from the one proposed by Vlachos \textit{et al.} \cite{vlachos} in two ways: (1) we have integrated a new temporal constraint on $\delta$ to better control how far in time we can go in order to match points, starting from the end of the subsequences -- $\left|N-n-M+m\right|\leq\delta$; and (2) we have extended the similarity measure to the consideration of heterogeneous parameters. The constraint on values for similarity is based on the distance between points defined for each type of parameters in the previous paragraph \ref{ptsdist}. We have also defined a relevant $\epsilon$ threshold on these distances according to the parameter's type, considering that two values of a qualitative parameter are similar only if they are equal:\\
\begin{tabular}{ll}
        $\bullet$ Quantitative & $0<\epsilon<1$,\\
        $\bullet$ Ordered qualitative & $\epsilon = \frac{1}{v-1}$,\\
        $\bullet$ Unordered qualitative & $\epsilon = 1$.
\end{tabular}

The number of matching is normalized by the minimum length of the two trajectories, so that the similarity measure ranges from 0 to 1. Therefore the function $D_{\delta,\epsilon}(A,B)$ between the two trajectories $A$ and $B$ is defined as follows \cite{vlachos}:
        \[D_{\delta,\epsilon}(A,B) = 1 - \frac{LCSS_{\delta,\epsilon}(A,B)}{min(n,m)}.\]
$D_{\delta,\epsilon}(A,B)$ verifies the properties of a distance.

\subsection{Minimum distance}\label{mindist}

The minimum distance between time-series is a low approximation of the actual distance, which is interestingly used for getting a rough idea of the similarity between two sequences. Computing a minimum distance of zero between two subsequences means that they can be considered as quite similar. In our context, we use and interpret this information in terms of allowing for temporal aggregation of a sequence of vectors, when close to the corresponding ``mean vector sequence'' (see \ref{temporal_aggregation}). Chiu \textit{et al.} \cite{chiu} have defined a minimum distance to roughly compare discrete one-dimensional quantitative temporal sequences, in a purpose of classification. In our context, such a distance may also be of great interest to perform temporal aggregation, giving an idea of whether a subsequence can be approximated by its mean vector or not. We then need to extend the minimum distance of Chiu \textit{et al.} \cite{chiu} to allow for heterogeneous multivariate time-series.

The definition uses the values of breakpoints defining the discrete intervals of values. 
Let $B = \beta_1,...,\beta_{a-1}$ be the sorted list of breakpoints for a given quantitative parameter discretized in $a$ symbols $\alpha_1$,...,$\alpha_a$ ($\beta_0$ and $\beta_a$ are defined as $-\infty$ et $+\infty$ respectively). 
A sequence $C = c_1,...,c_n$ of length $n$ can be transformed into a symbolic representation as a word $\hat{C} = \hat{c}_1,...,\hat{c}_\omega$ where $\hat{c}_i = \alpha_j$ iff $\beta_{j-1} \leq c_i < \beta_{j}$. Using the principle of Euclidean distance, the minimum distance between the original time-series $Q$ and $C$ of two words $\hat{Q}$ and $\hat{C}$, so called $mindist(\hat{Q},\hat{C})$, is defined by the following equation \cite{chiu}: 
\[ mindist(\hat{Q},\hat{C}) = \sqrt{\frac{n}{\omega}} \sqrt{\sum_{i=1}^\omega\left(d(\hat{q}_i,\hat{c}_i)\right)^2}.
\]
The distance function $d(\alpha_i,\alpha_j)$ between two symbols $\alpha_i$ and $\alpha_j$ of a given ordered alphabet corresponding to discretization intervals, $1\leq i,j\leq a$, is defined using the values of the corresponding breakpoints, as follows \cite{chiu}:
\begin{equation}\label{quant}
        d(\alpha_i,\alpha_j) = \left\{
        \begin{tabular}{ll}
        0 & \mbox{if $\left|i-j\right|\leq1$ },\\
        $\beta_{max(i,j)-1} - \beta_{min(i,j)}$ & \mbox{otherwise}.
        \end{tabular}
        \right.
\end{equation}
The implementation of a lookup table used to define the distance between words made up of such symbols is illustrated in table \ref{symb_dist}.
\begin{table}[h]
\begin{center}
\begin{tabular}{ccccc}
&$\alpha_1$&$\alpha_2$&$\alpha_3$&$\alpha_4$\\
\cline{2-5}
$\alpha_1$&0&0&0.25&0.57\\
\cline{2-5}
$\alpha_2$&0&0&0&0.32\\
\cline{2-5}
$\alpha_3$&0.25&0&0&0\\
\cline{2-5}
$\alpha_4$&0.57&0.32&0&0\\
\cline{2-5}
\end{tabular}
\end{center}
\caption{\small \textbf{Distance between discrete symbols representing time-series.}}\label{symb_dist}
\small Lookup table used to compute the minimum distance between two words for an alphabet of cardinality $4$, $\alpha_1,...\alpha_4$, defined by discretization using the breakpoints $\beta_1=0.12$, $\beta_2=0.37$, and $\beta_3=0.69$. The distance between two symbols can be read off by examining the corresponding row and column. For instance $d(\alpha_1,\alpha_2)=0$ and $d(\alpha_1,\alpha_3)=0.25$.
\end{table}

To extend this notion of minimum distance to multidimensional heterogeneous time-series, we use the distance function between points defined in section \ref{ptsdist} for qualitative parameters. The implementation of a lookup table used to define the distance between sequence of qualitative symbols is illustrated respectively in tables \ref{symb_unord_dist} and \ref{symb_ord_dist}.
\begin{table}[h]
\begin{center}
\begin{tabular}{ccccc}
&$\alpha_1$&$\alpha_2$&$\alpha_3$&$\alpha_4$\\
\cline{2-5}
$\alpha_1$&0&1&1&1\\
\cline{2-5}
$\alpha_2$&1&0&1&1\\
\cline{2-5}
$\alpha_3$&1&1&0&1\\
\cline{2-5}
$\alpha_4$&1&1&1&0\\
\cline{2-5}
\end{tabular}
\end{center}
\caption{\small \textbf{Distance between symbols from an unordered alphabet.}}\label{symb_unord_dist}
\small Lookup table used to compute the minimum distance between two words for an unordered alphabet of cardinality $4$, $\alpha_1,...\alpha_4$.
\end{table}
\begin{table}[h]
\begin{center}
\begin{tabular}{cccc}
&$\alpha_1$&$\alpha_2$&$\alpha_3$\\
\cline{2-4}
$\alpha_1$&0&0.5&1\\
\cline{2-4}
$\alpha_2$&0.5&0&0.5\\
\cline{2-4}
$\alpha_3$&1&0.5&0\\
\cline{2-4}
\end{tabular}
\end{center}
\caption{\small \textbf{Distance between symbols from an ordered alphabet.}}\label{symb_ord_dist}
\small Lookup table used to compute the minimum distance between two words for an ordered alphabet of cardinality $3$, $\alpha_1<\alpha_2<\alpha_3$.
\end{table}
\\Let $C = ((c_{1,1},...,c_{1,p}),...,(c_{n,1},...,c_{n,p}))$ be a $p$-dimensional heterogeneous time-series represented by the sequence of symbols $\hat{C} = ((\hat{c}_{1,1},...,\hat{c}_{1,p}),...,(\hat{c}_{n,1},...,\hat{c}_{n,p}))$. Using the relevant function $d(\hat{q}_{i,j},\hat{c}_{i,j})$ according to the type of component $j$ -- equation (\ref{dist_qual_ord}), (\ref{dist_qual_unord}), or (\ref{quant}) using the normalized quantitative values -- the minimum distance between two original $p$-dimensional time-series $Q$ and $C$ represented as $\hat{Q}$ and $\hat{C}$, so called $mindist(\hat{Q},\hat{C})$, is then re-defined by the following equation:
\[ mindist(\hat{Q},\hat{C}) = \sqrt{\frac{n}{\omega}} \sqrt{\sum_{i=1}^\omega\left(\sum_{j=1}^p\left(d(\hat{q}_{i,j},\hat{c}_{i,j})\right)^2\right)}.
\]

\section{Proposed approach for pattern extraction}\label{patterns_extraction}

\begin{figure}[t]
\begin{center}
\includegraphics[width=3in]{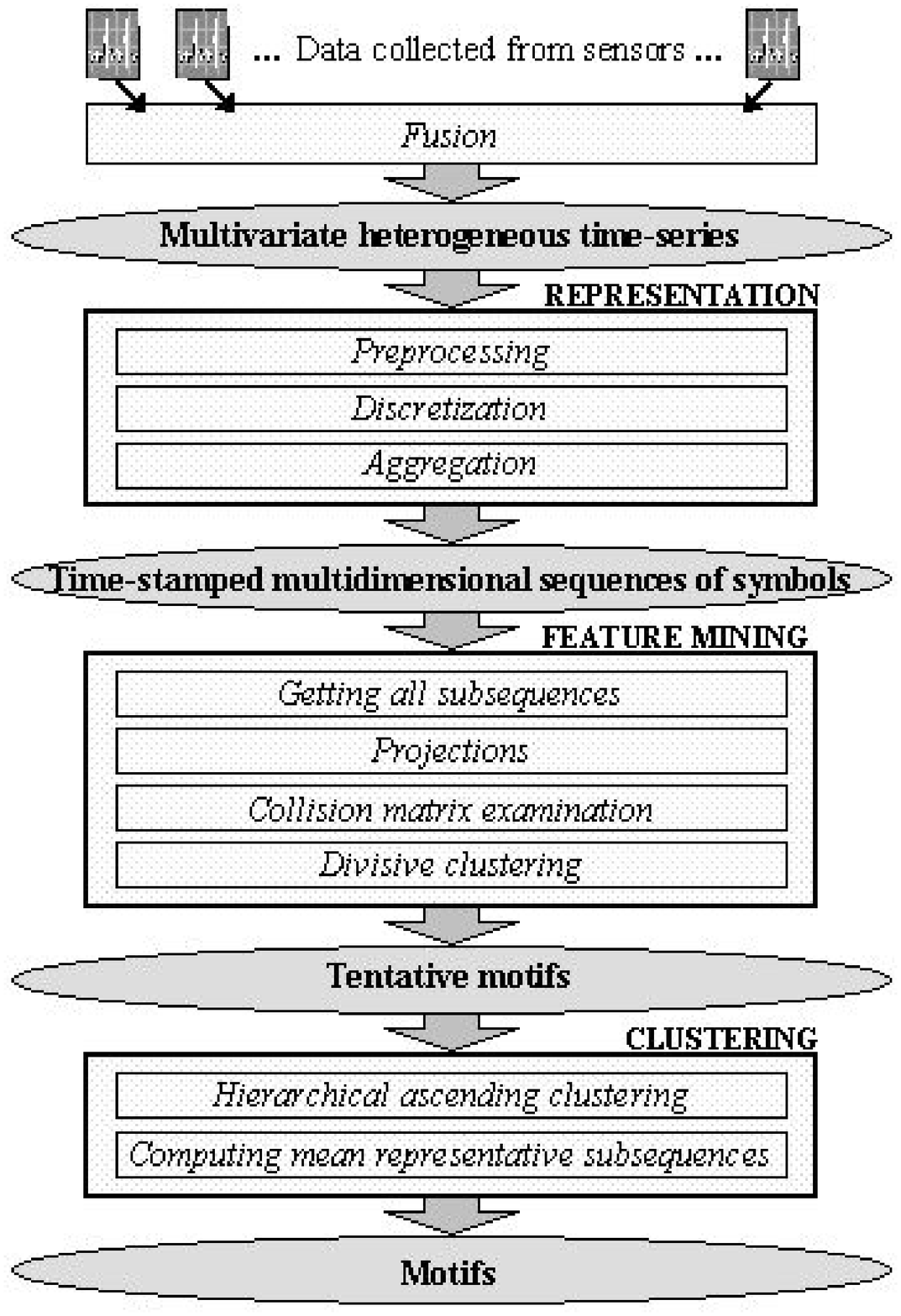}
\end{center}
\caption{\small \textbf{Successive steps of motifs extraction from data collected from sensors,}}\label{fig_extraction_steps}
\centering \small including: (1) Representation, (2) Feature mining, and (3) Clustering.
\end{figure}

In that section we describe the proposed approach for recurrent pattern extraction. The schema of figure \ref{fig_extraction_steps} summarize the successive steps identified in section \ref{methodology} for an unsupervised learning of meaningful ``high-level patterns'' from heterogeneous multivariate time-series, detailed in the following paragraphs: (\S \ref{representation}) \textbf{Representation} of time-series, (\S \ref{tent_motifs_discovery}) \textbf{Feature mining} for tentative motifs discovery, and (\S \ref{time_series_motifs}) \textbf{Clustering} for time-series motifs final identification.

\subsection{Representation of time-series}\label{representation}

Time-series representation is really important because of the difficulty of directly manipulating continuous, and especially heterogeneous, high-dimensional and possibly noisy data in an efficient way. Defining a suitable representation aims at reducing feature space dimension in order to get an efficient feature mining for running the learning task.
Many time-series representation have been introduced, including the Discrete Fourier Transform (DFT), the Discrete Wavelet Transform (DWT), Piecewise Linear and Piecewise Constant models (PAA, APCA), Singular Value Decomposition (SVD) (see \cite{antunes} for an overview).

Since we are looking for ``high-level patterns'' (see \S \ref{guidelines}) within the time-series -- that is, patterns corresponding to usual activities of a person at home -- our purpose of representation is to highlight the global trends within the data, while removing minor local variations. Our main concern is in fact robustness rather than accuracy of the extracted patterns. In that aim, we also try to restrict as much as possible the number of parameters involved in the process. Defining the step of representation is then guided by this purpose of getting a long-term, simple, and meaningful view of the time-series, which corresponds in fact to a step of \textbf{abstraction}.

\begin{figure}[t]
\begin{center}
\includegraphics[width=2.9in]{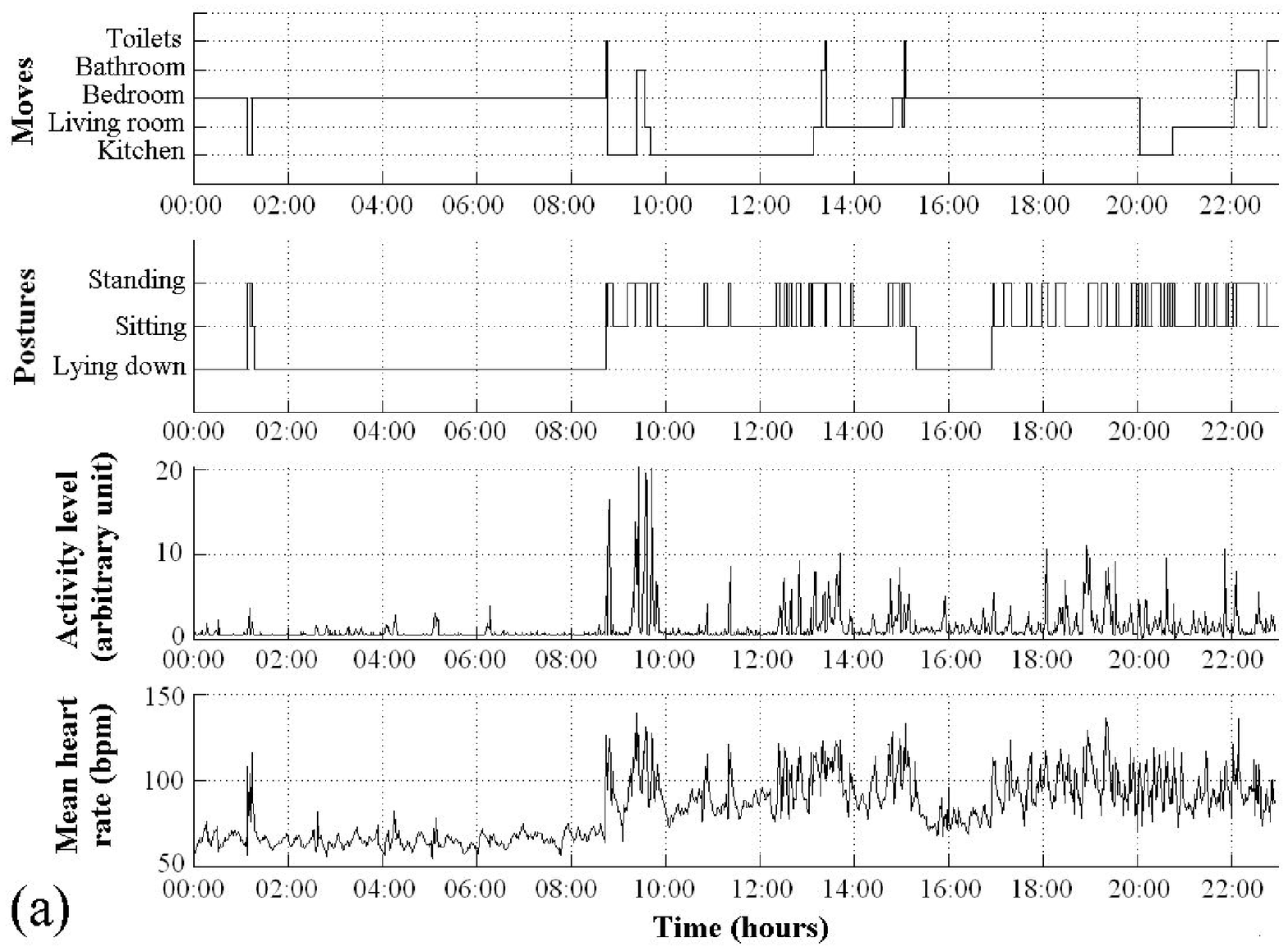} \includegraphics[width=2.9in]{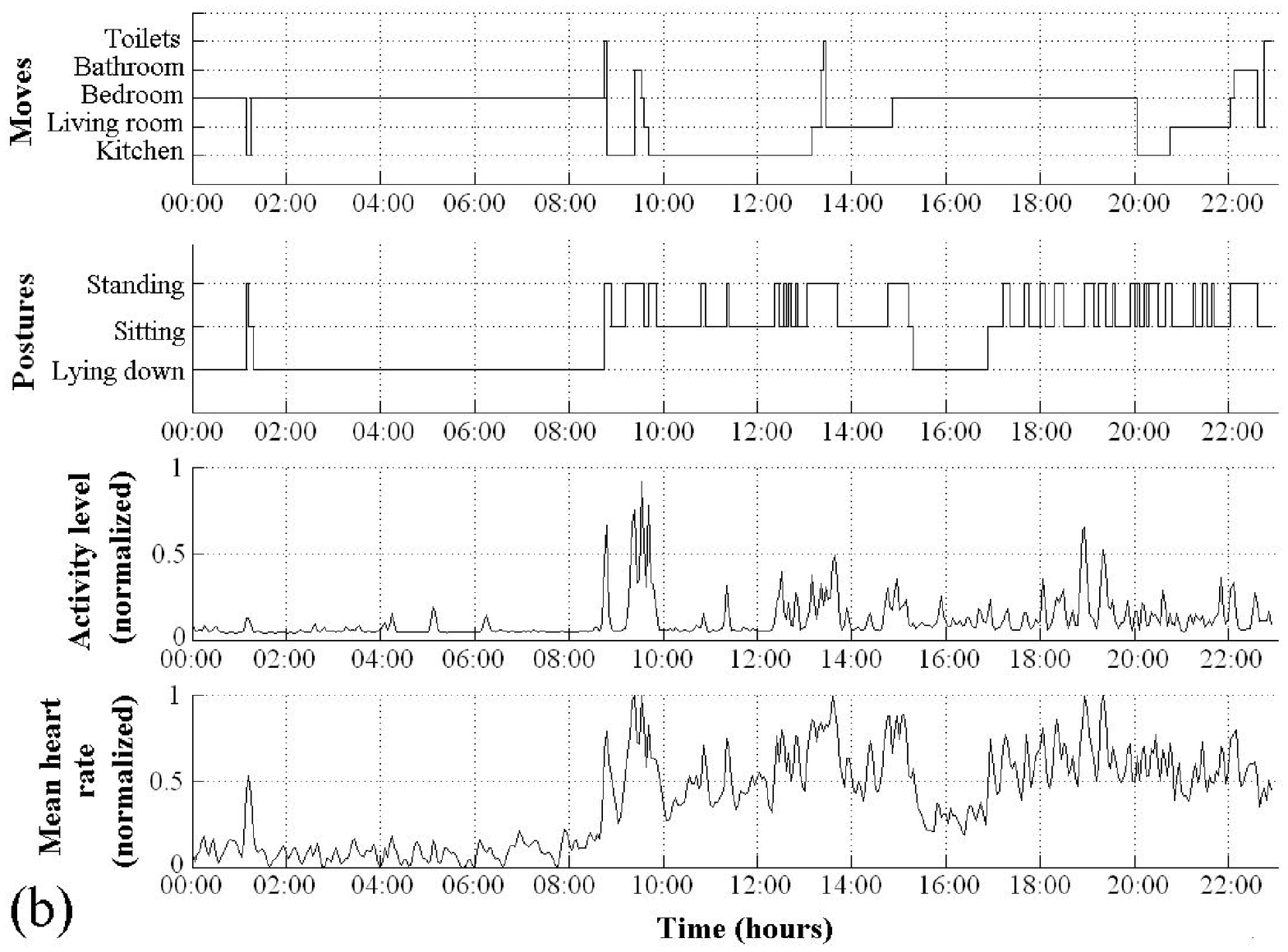} \\
\includegraphics[width=2.9in]{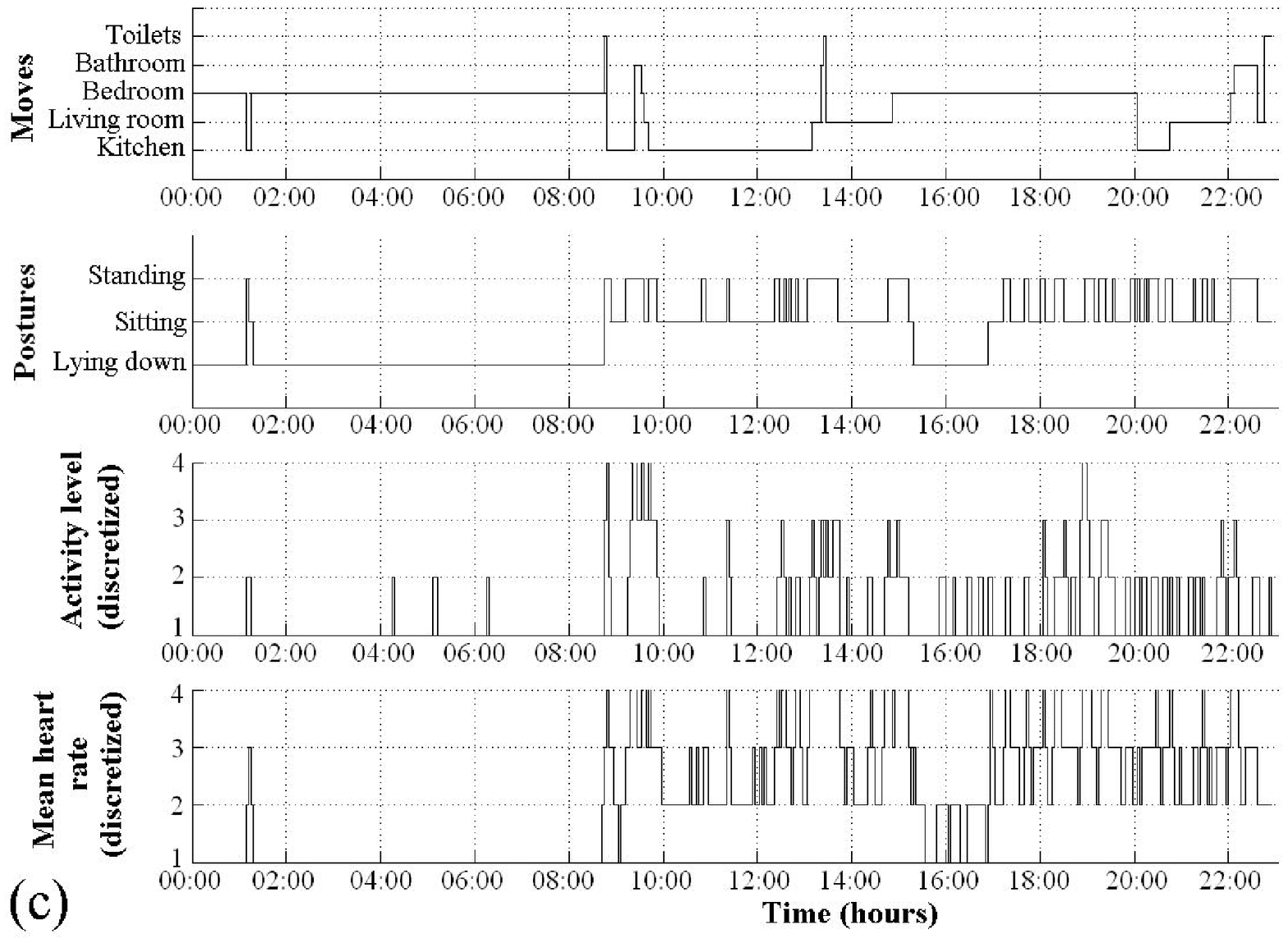} \includegraphics[width=2.9in]{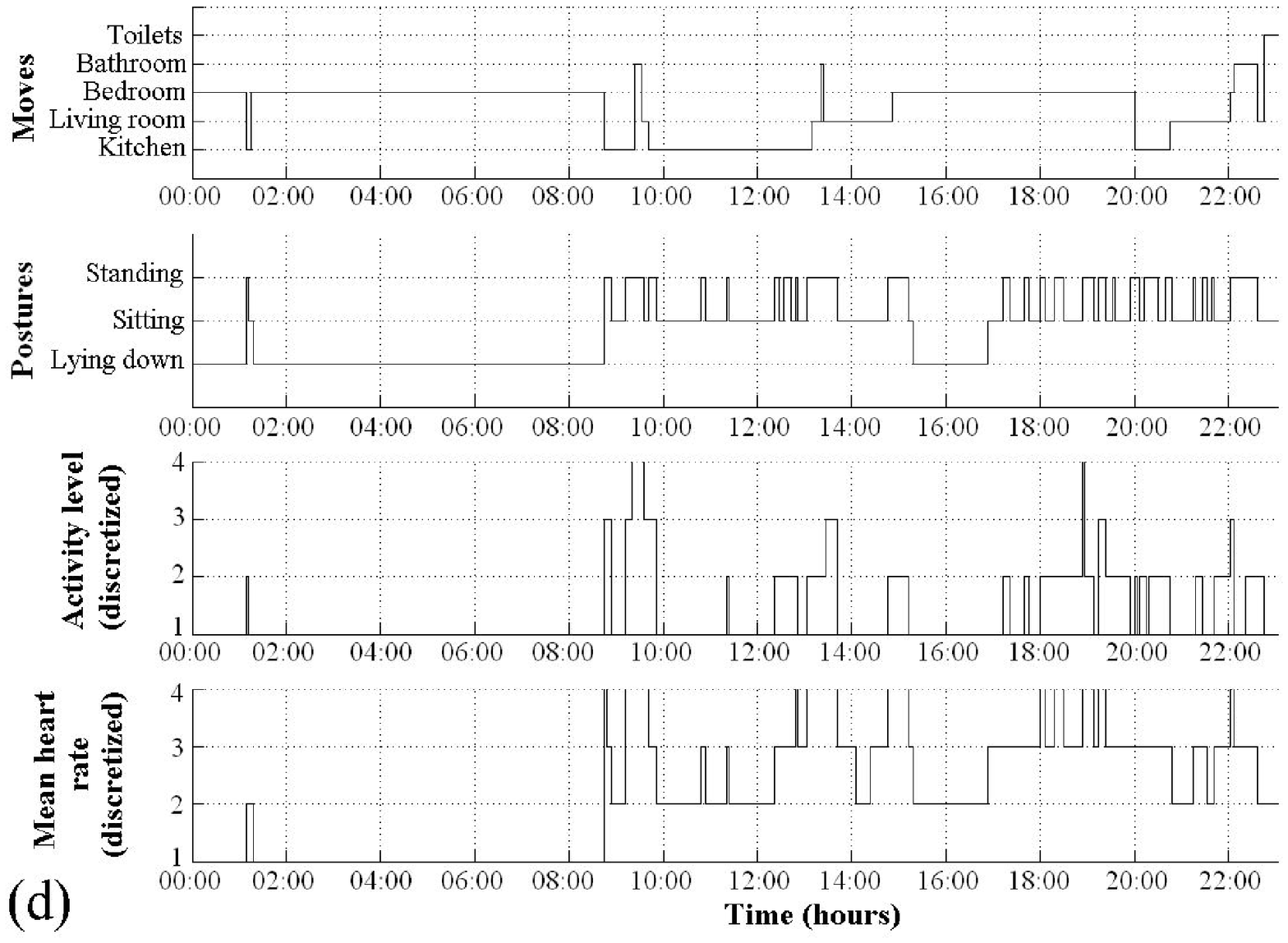}
\end{center}
\caption{\small \textbf{Abstraction steps performed before analyzing time-series.}}\label{fig_timeseries_data}
\small Graphs representing some sequences of 4-dimensional time-series for a person monitored at home over one day, including for each graph the following parameters (from top to bottom): (1) moves, (2) postures, (3) activity levels, and (4) mean heart rate.

From left to right, and from top to bottom, the graphs represent: (a) \textbf{Raw data}, produced by a simulation process; (b) \textbf{Preprocessed data}, that raw data smoothed by a mean filter, following by temporal reduction, and normalization; (c) \textbf{Discretized data}, the discretized intervals being defined by the k-means technique, and (d) \textbf{Aggregated data}.
\end{figure}

We perform time-series abstraction in three steps to get a concise representation of the heterogeneous multivariate time-series under study: (1) preprocessing, (2) discretization, and (3) temporal aggregation. Preprocessing the time-series includes filtering, temporal reduction and alignment . Although well known and usual when analyzing data sets, this step is also really important because it at least partly governs the level of details of the analysis. The next subsections detail the two following steps of discretization and temporal aggregation. The figure \ref{fig_timeseries_data} illustrates the results of each of these steps from a sample sequence simulated in the contexte of home health telecare. 

\subsubsection{Discretization}\label{discretization}

Dealing with heterogeneous time-series, using a symbolic or discrete representation of the lower-level data is interesting to build homogeneous data sets for feature mining. This requires the \textbf{discretization} of the continuous components of time-series. Several methods have already been experimented like Piecewise Aggregate Approximation (PAA) to produce symbols of equiprobability \cite{chiu, lin}, clustering using k-means \cite{das_b} or Dynamic Local K-means (DLK) -- DLK learns the number of the classes with subject to the constraint that the variance of each class is less than a given sigma-zero \cite{hong_b}. Given that equiprobability in symbols is not necessarily relevant considering monitoring purposes -- unusual values need to be distinguishable from usual ones -- we use the standard k-means technique on experimental data sets to define the discretization intervals for quantitative parameters.

\subsubsection{Temporal aggregation}\label{temporal_aggregation}

Once we get discrete time-series, a further step in reducing the feature space dimension is to perform \textbf{temporal aggregation}, where aggregate vectors -- either called \emph{symbols} -- are computed over time-line partitions. The main interest is to get a concise representation of time-series, allowing stretching in time between subsequences represented by a same number of symbols, and possibly similar in terms of this aggregated representation.
In general, temporal grouping is done by two types of partitioning \cite{moon}: (1) span grouping, based on a defined length in time, and (2) instant grouping, which depends on the data stored. Various techniques have already been proposed and applied to issues where several time-series of same parameters are recorded during overlapping time-intervals \cite{meratnia, moon}. The aim is then to summarize the time variations over only one possibly multidimensional sequence of values. 

In our context, the issue is quite simpler because it only aims at partitioning one multidimensional sequence of discrete values into a time-stamped sequence of vectors which summarizes the global trends of variation. \textbf{Span aggregation} may be performed while pre-processing the time-series using sliding windows of fixed-length to mean the data and possibly reduce the sampling rate in the same time. The choice of sampling rate is important to determine the precision and thus the interesting level of details of time-series. A more challenging and also common issue is \textbf{instant aggregation}, depending on the variation of the values in time. Since our interest is in observing global trends in the time-series, we need to compute aggregate vectors within time-intervals where there are no significant variations in the multidimensional values (that is, vectors). We use a technique based on a distance threshold using an extension of the minimum distance between time-series proposed in \cite{chiu}. The great interest in using a minimum distance (see \S \ref{mindist}) is the ease of getting a low approximation of the actual distance, which can be intuitively interpreted in terms of relative global trends of variation between time-series. In order that the aggregation of a sequence into one time-stamped vector is performed only for successive similar vectors, we decide that the minimum distance between the original and aggregated sequences must not be over zero for allowing aggregation. Considering the heterogeneous case, a minimum distance of zero means values of quantitative parameters are similar along time -- that is within adjacent discretized intervals -- and values of qualitative parameters are the same. 

The aggregate vector $\widehat{aggr(C)}$ of a $p$-dimensional sequence of length $n$,
\[C = ((c_{1,1},...,c_{1,p}),...,(c_{n,1},...,c_{n,p})),
\]
is defined as the discretized mean value computed along time for each component:
\[\widehat{aggr(C)}=(\widehat{mean\left(c_{1,1},...,c_{n,1}\right)},...,\widehat{mean\left(c_{1,p},...,c_{n,p}\right))}.
\]
The mean value of a symbolic sequence corresponds to the most represented symbol within the whole subsequence. The following equation express the condition required for aggregating the vectors of a subsequence $C$:
        \[mindist(\hat{C},\widehat{AGGR(C)})=0.
\]
$\hat{C}$ is the discretized sequence of $C$; $\widehat{AGGR(C)}$ is a sequence of same length $n$ made up of the repeated aggregate vector, 
        \[\widehat{AGGR(C)}=(\widehat{aggr(C)},...,\widehat{aggr(C)});
\]
and $mindist()$ is the minimum distance whose definition is extended from the one proposed by from Chiu \textit{et al.} \cite{chiu}.
Starting from the first point of the sequence, we look for the longest time-intervals where temporal aggregation is allowed according to the previous definitions. At the end, the original time-series is then represented by a sequence of multidimensional vectors, either called symbols, each of them lasting for a specific duration.

\subsection{Feature mining: tentative motifs discovery}\label{tent_motifs_discovery}

\begin{figure}[t]
        \begin{center}
                \includegraphics[width=\textwidth]{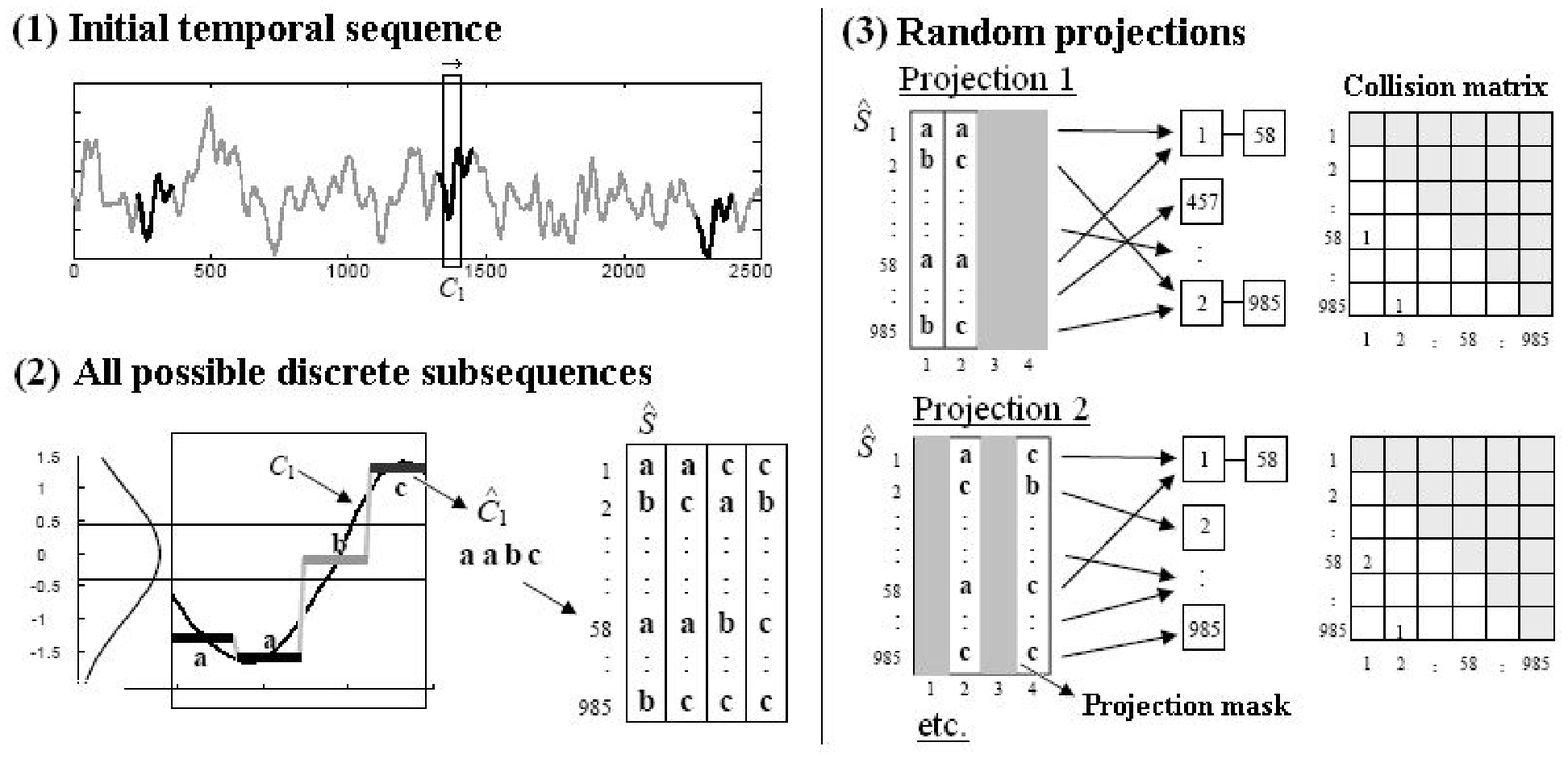}
        \end{center}
        \caption{\small \textbf{Principle of using the \emph{projection} algorithm as experimented in \cite{chiu}.}}\label{fig_projections_chiu}
\small The successive steps are as follows :
\begin{liste_par_arabic}
\item A sliding window is defined to extract subsequences $C_1$ from the initial temporal sequence ;
\item Each sequence $C_1$ is converted into its discrete representation $\hat{C}_1$ and placed into matrix $\hat{S}$ ;
\item A mask is randomly chosen, so that only part of the discrete values were used to project the matrix $\hat{S}$ into buckets. Collisions are recorded by incrementing the appropriate location in the collision matrix.
\end{liste_par_arabic}
\end{figure}

Feature mining aims at selecting the most relevant features to feed into a learning task, so that we reduce the size of the feature space. In the context of time-series motifs discovery, the purpose is to extract the most relevant subsequences -- the \emph{tentative motifs} -- used as input for the final identification and classification of frequent patterns -- the \emph{motifs}.

Several papers dealing with motifs extraction use a kind of feature mining step to first select the potential location of frequent patterns within the time-series, and then refine the motifs identification. For instance, Hong \textit{et al.} \cite{hong} have trained recurrent neural nets to extract temporal patterns candidates. They correspond to subsequences where the trained network can continuously give out correct one-step prediction. 
In that paper we extend the probabilistic approach experimented in \cite{chiu} for tentative motifs extraction, and illustrated on figure \ref{fig_projections_chiu}. Time-series motifs candidates -- that is \emph{tentative motifs} -- are identified from random projections of all the possible subsequences extracted using a sliding window from the original and symbolized time-series. The tentative motifs correspond to subsequences that are often hashed into the same bucket using a mask randomly chosen. Each step of projection increases the counts in a \emph{collision matrix}, a square matrix whose size corresponds to the number of all possible subsequences of a predefined length. A large value of collisions is a strong indicator of two similar subsequences, that is good candidates for motifs. Since time-series are discretized into symbols of constant frequency, this method does not allow for stretching in time between motifs instances. Moreover, Chiu \textit{et al.} have implemented the \emph{projection} algorithm \cite{buhler} only for one-dimensional real time-series.

Actually, the \emph{projection} algorithm is interesting because of its ability to roughly identify possible instances of motifs in time-series, allowing some noise and imprecision in the discrete sequences representing the original time-series. In our complex issue involving several levels of details from raw data to decision, this algorithm then acts as a stage of feature meaning to extract the best subsequences that are candidates to motifs from the original time-series represented as a discrete sequence of symbols.
We then extend this approach to deal with heterogeneous multivariate time-series. The context of using the \emph{projection} algorithm is also changed so that we can discover motifs whose instances are of different lengths. Using as input discrete sequences obtained with representation techniques including temporal aggregation of symbols (see \S \ref{temporal_aggregation}) addresses this issue.

Using the \emph{projection} algorithm allows to extract subsequences representative of frequent patterns within discrete time-series. This criterion is however not enough to deal with feature mining -- that is tentative motifs extraction. According to Lesh \textit{et al.} \cite{lesh}, the criteria for selecting features might depend on the domain and the classifier being use. However, they believe that the following domain- and classifier-independent heuristics are useful for selecting sequences to serve features: (1) Features should be frequent, (2) Features should be distinctive of at least one class, and (3) Feature sets should not contain redundant features.
\renewcommand{\labelenumi}{(\theenumi)}
\begin{enumerate}
        \item The first heuristic is clearly insured because subsequences extracted using the \emph{projection} algorithm at least partially matches another subsequence due to their extraction from the collision matrix produced by the successive projections.
        \item The second one cannot be encoded directly from projections because this approach to pattern extraction is rough and unsupervised. Having features distinctive of at least one class is then ensured by three additional steps when examining the collision matrix. First, because a large value of collisions is only a strong indicator of similar subsequences, we go back to the original, preprocessed, time-series to refine the comparison between pairs of subsequences. Second, because a motifs' instances may be of different lengths regarding the number of corresponding discrete symbols, each instance may be represented by several successive \emph{basic} subsequences. Consequently, we propose to extend matching pairs of subsequences while examining the collision matrix, as long as they are similar, in order to match the whole patterns. Third, we add a constraint on the minimum duration of so extracted subsequences to insure that features are relevant in terms of the person's behavioral profile, and consequently probably in terms of being an instance of a motif.
        \item At last, the third heuristic requires synthesizing the set of subsequences identified from the previous steps in order to get a set of non-overlapping features -- that is, subsequences -- that are the most representative of each group of overlapping subsequences. The need for synthesizing the results of collision matrix examination has not been put forward in related papers. This stage then ends feature mining, that is the tentative motifs extraction, with respect to the third heuristics of Lesh \textit{et al.} \cite{lesh}.
\end{enumerate}
\renewcommand{\labelenumi}{\theenumi.}

\noindent Feature mining then includes three main steps, detailed in the next subsections:
\begin{enumerate}
        \item \textbf{Time-series random projections}, once represented using discretization and aggregation techniques,
        \item \textbf{Collision matrix examination} to extract frequent and relevant subsequences in terms of identifying motifs, and
        \item \textbf{Tentative motifs extraction} by identifying the most relevant, non-overlapping, subsequences from the previous set. 
\end{enumerate}

\begin{figure}[t]
        \begin{center}
                \includegraphics[width=9cm]{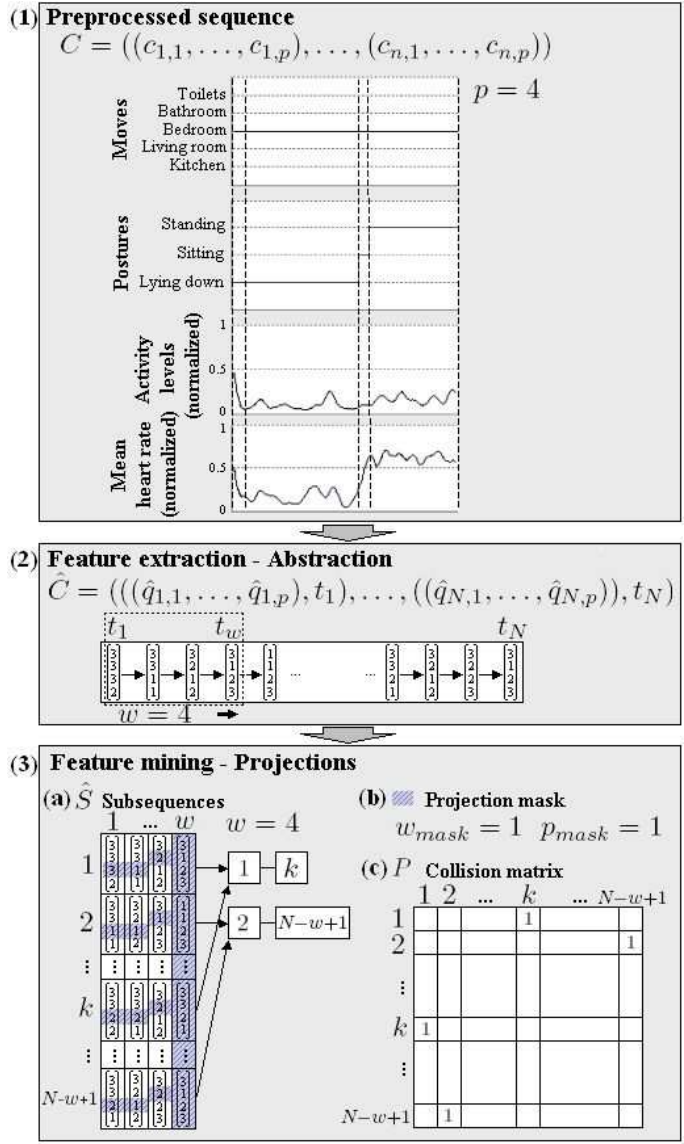}
        \end{center}
        \caption{\small \textbf{Principle of feature mining, once original time-series are represented.}}\label{fig_projections}
\end{figure}

\subsubsection{Time-series random projections}

Time-series random projections produce a \emph{collisions matrix} recording integers representative of the number of matches between all the possible subsequences extracted from the original sequence. A large value in a cell is not the guarantee of the existence of a corresponding motif, but it is a strong indicator \cite{chiu}.
The diagram of figure \ref{fig_projections} sums up random projections principle, according to the following steps:
\begin{liste_par_arabic}
\item \textbf{Preprocessing.} Let $C$ be a $p$-dimensional sequence of $n$ values recorded regularly over time:
\[C = ((c_{1,1},...,c_{1,p}),...,(c_{n,1},...,c_{n,p})).\]

\item \textbf{Abstraction.} Time-series are first represented using the techniques presented in section \ref{representation} -- preprocessing, discretization, and temporal aggregation. Sequence $C$ is then represented by a time-stamped sequence of $N$ symbols -- the $p$-dimensional vectors $\left(\hat{q}_{i,1},...,\hat{q}_{i,p}\right)$, $1 \leq i \leq N$, $N \leq n$:
\[\hat{C} = (((\hat{q}_{1,1},...,\hat{q}_{1,p}),t_1),...,(( \hat{q}_{N,1},...,\hat{q}_{N,p})),t_N),\]
where $(t_1,...,t_N)$ are the ordered instants of symbols occurrence along time.

\item \textbf{Random projections}

\begin{liste_par_alph}
\item \textbf{Basic subsequences.} Random projections are performed from so called \emph{basic subsequences} of a specified length $w$, extracted from the original sequence using a sliding window of size $w$. This produce a matrix $\hat{S}$ of size $(N-w+1)\times w$.

\item \textbf{Projection mask.} We randomly select $w_{mask}$ columns of $\hat{S}$ to act as a mask. In a $p$-dimensional context we need a random mask of size $w_{mask}\times p_{mask}$, where $w_{mask}$ and $p_{mask}$ are integers such as $0\leq w_{mask} \leq w$ and $0\leq p_{mask} \leq p$. For example in figure \ref{fig_projections}, where $w=4$ and $p=4$, we have randomly selected column number 4 ($w_{mask}=1$) to act as a mask on symbols, and parameter number 3 for symbols 1 and 2, such as parameter number 2 for symbol 3 ($p_{mask}=1$).  

\item \textbf{Collision matrix.} The $(N-w+1)$ words in the $\hat{S}$ matrix are hashed into buckets based on their non-masked values. In the sample of figure \ref{fig_projections}, all possible couple of subsequences are compared based on their $1^{nd}$, $2^{nd}$, and $3^{nd}$ symbols, considering the $1^{st}$, $2^{nd}$ and $4^{th}$ parameters for symbols 1 and 2, and the $1^{st}$, $3^{rd}$ and $4^{th}$ parameters for symbol 3. This produces the \emph{collision matrix} $P$ of size $(N-w+1)\times(N-w+1)$, built as follows: if two words corresponding to $p$-dimensional subsequences $i$ and $j$ are hashed to the same bucket, we increase the count of cell $(i,j)$ in $P$, previously initialized to all zeros -- $P(i,j) = P(i,j) + 1$.

\end{liste_par_alph}

We need to repeat the two last steps (b) and (c) an appropriate number of times so that the collision numbers are statistically significant. The relevance of the collision numbers also depends on the parameters being well selected according to the purpose of motifs extraction. 

\end{liste_par_arabic}

\noindent The key parameters of projection are as follows: (1) number of vectors per subsequence ($w$), (2) number of vectors defining the first dimension of the projection mask ($w_{mask}$), (3) number of parameters defining the second dimension of the projection mask ($p_{mask}$), (4) number of projections performed ($proj$).

\subsubsection{Collision matrix examination}

Once the collision matrix is significantly filled in, we examine iteratively its values from the largest one to find promising candidates for motifs extraction. We stop the examination when the next value not already examined, and not within the scope of previously reported tentative motifs, is lower than a predefined threshold defining the minimum ``large enough'' number of collisions, so called the \textbf{\emph{collision threshold}}. A large value of collisions is only a strong indicator of two similar subsequences, and we go back to the original data to possibly confirm we met a \emph{tentative motif}. The comparison of the original, preprocessed, subsequences corresponding to tentative motifs is performed from the similarity measure defined in section \ref{similarity} for heterogeneous multivariate time-series. A threshold on this measure, so called the \textbf{\emph{distance threshold}}, is used to decide whether or not two subsequences can be considered similar enough so that they define tentative motifs.

In order to look for the \emph{whole} and significative patterns, independently of the number $w$ of symbols defining each \emph{basic subsequence} used as input for projections, we perform ``pattern growing'' -- as called in \cite{hong} in another methodological context -- while examining the collisions matrix. Considering a pair of similar discrete \emph{basic} subsequences in terms of collision number and actual distance, we define ``pattern growing'' as the consideration of extended subsequences including the basic ones. We try to extend them on their right and left sides -- that is before and after the respectively first and last symbols in time -- while (1) the numbers of collisions corresponding to the extended area is still large enough -- that is over the \textbf{\emph{collision threshold}}, and (2) the similarity between the extended original subsequences does not overpass the \textbf{\emph{distance threshold}}. 

\begin{figure}[t]
        \begin{center}
                \includegraphics[width=12cm]{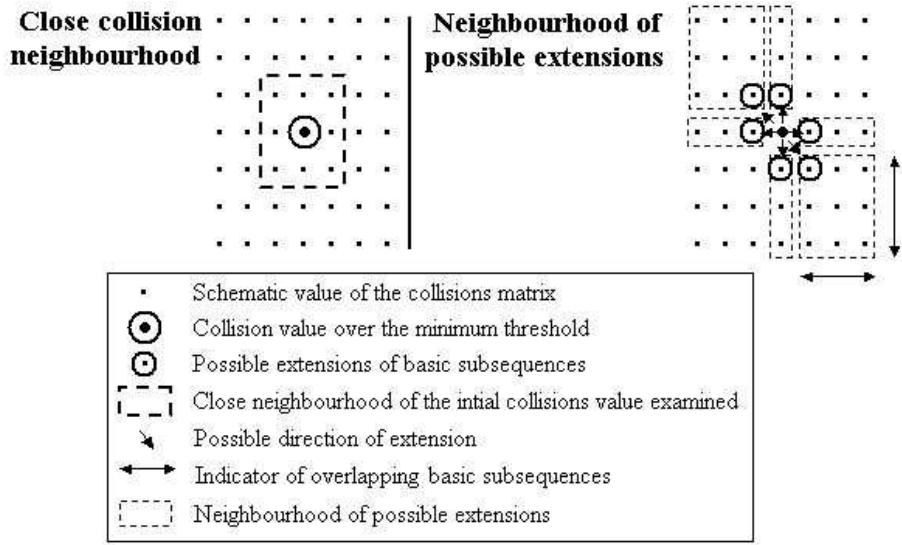}
        \end{center}
        \caption{\small \textbf{Neighbourhoods of collisions used to identify and extend basic recurrent subsequences from the collision matrix.}}\label{fig_ext_neigh}
\end{figure}

Moreover, we define some \emph{neighbourhoods of collisions} to allow for noise between recurrent subsequences and imprecision in the abstraction step. Indeed, in that noisy and imprecise context, all basic subsequences that made up a whole motif instance do not generate collision values over the predefined threshold when compared to other instances. An adaptive algorithm is then defined from the observation of a high collisions number, as illustrated on figure \ref{fig_ext_neigh} : (1) A finest identification of recurrent basic subsequences is performed by finding out the lowest actual distance, under the maximum threshold, between pairs of basic subsequences in a \emph{close collision neighbourhood} that verifies the minimum collisions criteria ; (2) Pattern growing is then performed when the collision criteria is verified in the \emph{neighbourhood of possible extensions} and the actual distance between the extended subsequences is under the maximum threshold.

We then produce a group of subsequences of different lengths in terms of symbols and/ or number of points regarding the original subsequence.

\subsubsection{Tentative motifs extraction}

Feature mining aims at identifying non-redundant features \cite{lesh}, that is a set of non-overlapping subsequences from the whole original sequence that are the most appropriate to motifs extraction. 
However, the group of subsequences previously extracted from the collision matrix examination may contain overlapping ones because they are extracted by pairs regarding an area of high values within the collisions matrix. We then need to identify relevant groups of subsequences that are well-separated in time, so that we can define at last the tentative motifs, one corresponding to each group.

Considering a group of $k$ subsequences, one overlapping all the other ones of the group, and each of them ranging from indexes $t_{j,1}$ to $t_{j,n_j}$ ($1\leq j \leq k$, $n_j>1$), where $t_{j,1}<t_{j,n_j}$ regarding the original sequence, the \emph{tentative motif} representative of this group is defined by the subsequence ranging from indexes $t^i$ to $t^f$, where: 
\begin{equation}
t^i = min\left(\left\{t_{j,1}\right\}_{1\leq j\leq k}\right)
\mbox{ and }
t^f = max\left(\left\{t_{j,n_j}\right\}_{1\leq j\leq k}\right).
\end{equation}
The idea is indeed to consider the collision and distance thresholds as restrictive enough so that pattern identification and growing from the collision matrix examination only results in defining significant tentative motifs. However, a \emph{tentative motif} met several times as matching different subsequences may not be extended enough any time because of possible noisy data or imprecision in frequent subsequences identification. The largest subsequence regarding all overlapping subsequences must then be considered as the \emph{tentative motif}.

\begin{figure}[t]
\begin{center}
\includegraphics[width=10cm]{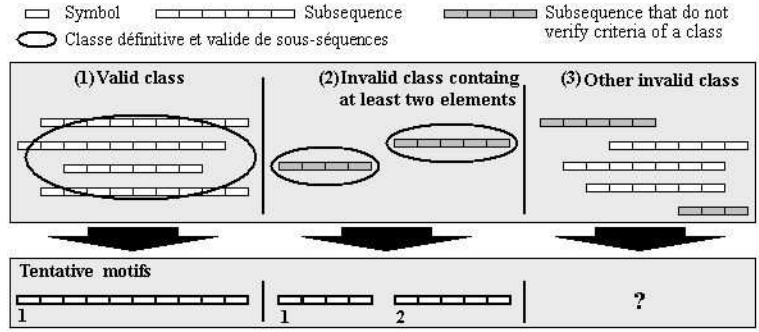}
\end{center}
\caption{\small \textbf{Illustration of possible cases when examining the validity of a class of tentative motifs.}}\label{motifs_cas_classes}
\end{figure}

However, frequent subsequences identified following the collision matrix examination may not be directly divided into well-separated groups of subsequences, where each group contains subsequences overlapping all the other ones. For instance, a subsequence corresponding to a large collision number regarding another subsequence might be hazardously too much extended in reference to the effective location of the corresponding motif. Regarding the group of overlapping subsequences containing this ``too long'' subsequence, removing this subsequence may result in the corresponding group being in fact possibly divided into two groups of well-separated subsequences. The best way of defining the corresponding \emph{tentative motif} is then to remove the consideration of the longest subsequence and to identify at the end two tentative motifs corresponding to each well-separated group of subsequences. Possible cases met when examining a group of tentative motifs are illustrated on figure \ref{motifs_cas_classes}.

Thus, defining the most relevant tentative motifs considering the results of the collision matrix examination raises an issue of clustering sets of subsequences. Within each set, one subsequence overlaps at least one other subsequence belonging to the same set, and no subsequence from any other set. Clustering is an unsupervised data analysis technique which searches to separate data items, having similar characteristics, in constituent groups. The most common clustering methods are partitioning, hierarchical agglomerative or hierarchical divisive ones \cite{chavent}. Agglomerative techniques start usually with single member clusters, whereas divisive methods begin with all cases in one large cluster. The divisive algorithm then subdivides it until some tests are satisfied. In theory, these could continue until there are $t$ clusters each containing one object, but in practice they usually stop at an earlier stage. Divisive methods are however more expensive that agglomerative ones.

\begin{figure}[t]
\begin{center}
\includegraphics[width=11cm]{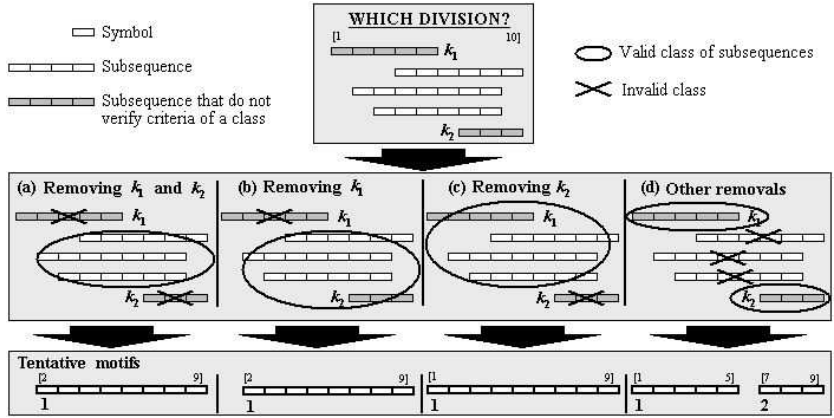}
\end{center}
\caption{\small \textbf{Illustration of possible divisions of an invalid class containing at least three subsequences.}}\label{motifs_cas_divisions}
\small A given class is considered as \emph{invalid} when at least two subsequences do not overlapp, so called $k_1$ and $k_2$.
\end{figure}

\begin{figure}[p]
\begin{center}
\includegraphics[width=14cm]{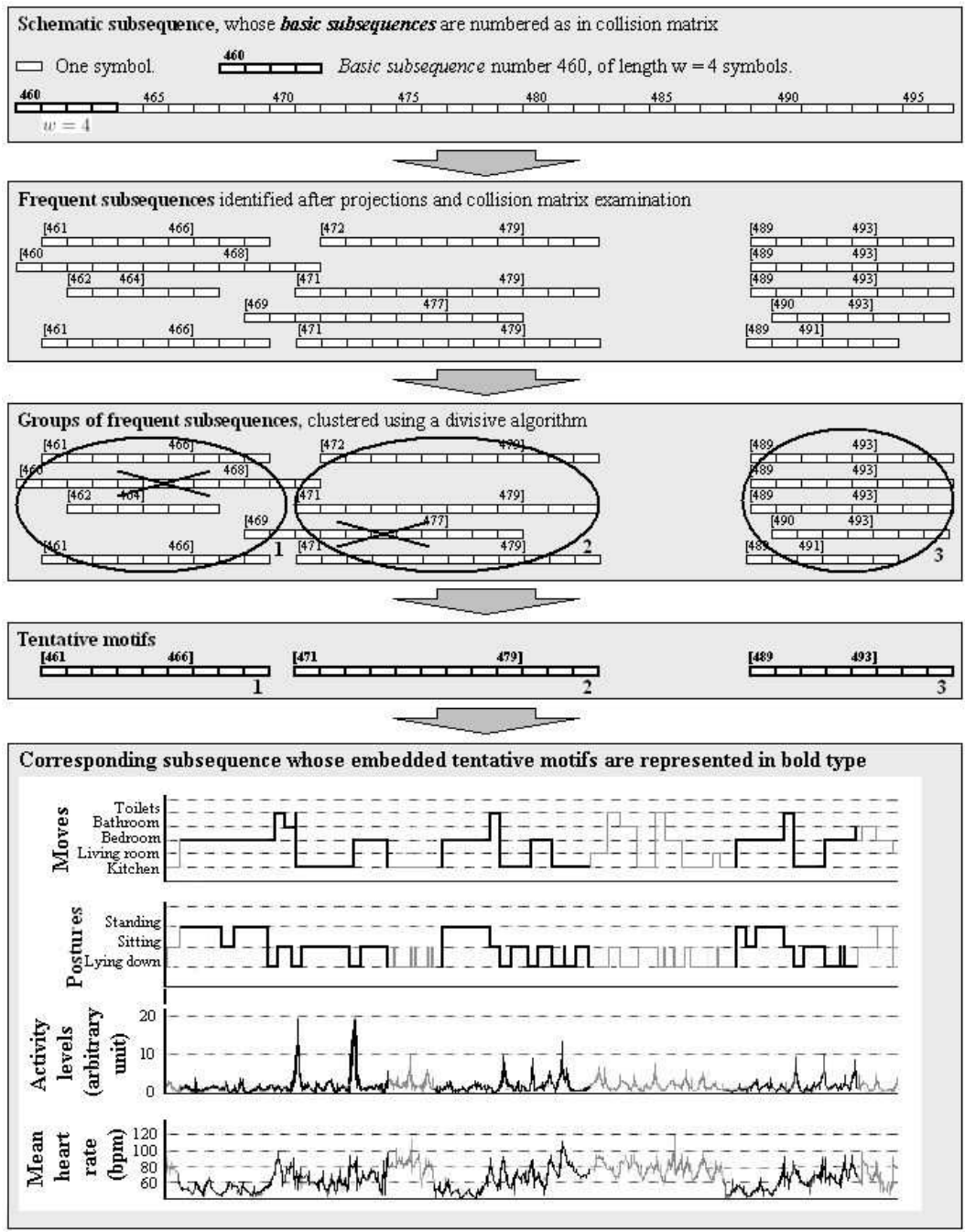}
\end{center}
\caption{\small \textbf{Tentative motifs extraction.}}\label{fig_tentative_motifs}
\small Successive steps to identify tentative motifs from the set of frequent subsequences identified after collision matrix examination, using a divisive clustering method.
\end{figure}

We use a hierarchical divisive clustering approach to decide about the best tentative motifs to be defined from a group of overlapping subsequences extracted at the end of examining the collision matrix. Criteria available to get clusters are indeed appropriate to that type of iterative algorithm, starting from one cluster that is gradually broken down into smaller and smaller clusters \cite{chavent}. At each iteration the next class to be divided is chosen, and the process is repeated until a given stop criteria is verified. The crucial elements to be defined are as follows: (1) the criteria to select the next class to be divided, (2) the method for dividing this class, and (3) the criteria to stop the successive divisions. 
The groups of subsequences considered as to be potentially divided are made up from the set of subsequences extracted when examining the collision matrix, as follows: any subsequence overlaps (a) at least one other subsequence from its group, and (b)no subsequence from any other group. The purpose of clustering the subsequences of each group is: any subsequence overlaps (a) all other subsequences of its group, and (b) no subsequence from any other group. The previous purpose defines the stop criteria for dividing the groups of subsequences. If a group of subsequences does not satisfy these constraints, that means some subsequences are not relevant to be considered and need to be removed from the set of frequent subsequences. 
\begin{enumerate}
        \item \textbf{Criteria of selecting the class to be divided.}
The selection of the groups to be next divided depends on the subsequences which does not overlap within the initial group. The criteria used to choose the ``best'' division to be performed -- that is also the ``best'' subsequence to be removed from the group so that all subsequences overlap each other -- is to end the divisions with the best representative groups of subsequences in terms of their well-representation of similar trends in the time-series. That is interpreted as removing the lower number of subsequences from the initial set. The ``best'' division is then determined \textit{a posteriori} considering all possible divisions. In order to prevent the algorithm from getting to an exponential running time, some optimization criteria are used to drive \textit{a priori} the selection of the best division, and to define what is an ``acceptable'' division -- that is an acceptable rate of subsequences removed. 
        \item \textbf{Method of dividing a class.}
At each step of dividing a class, three cases are possible, as follows: (1) either every subsequence overlaps all the other subsequences of the class, so that no division is required; (2) or there is only two subsequences that do not overlap within the class, so that the division consists in building two classes, containing one subsequence each; (3) or the class contains at least three subsequences, including at least two subsequences that do not overlap. In that third case, one ore more subsequences need to be removed from the class so that we can get to satisfy the criteria required to stop the divisions. Let $k_1$ and $k_2$ be two non-overlapping subsequences of a class. The division can be performed in the following manners (see figure \ref{motifs_cas_divisions}): (a) either removing $k_1$ and $k_2$, removing other subsequences from the group, (b) or removing $k_1$, (c) or removing $k_2$, (d) or keeping $k_1$ and $k_2$. Removing the consideration of one or more subsequences might entail that the original group is divided into  well-separated groups in time. We then apply the division algorithm on the corresponding new group(s) of subsequences.
        \item \textbf{Criteria to stop the successive divisions.}
We stop this recursive process of division when the purpose of clustering the subsequences of a group is reached, that is: any subsequence overlaps (a) all other subsequences of its group, and (b) no subsequence from any other group.
\end{enumerate}
Some results of identifying tentative motifs from the set of subsequences extracted from the collision matrix are presented on figure \ref{fig_tentative_motifs}.

\subsection{Clustering: time-series motifs identification}\label{time_series_motifs}

\begin{figure}[p]
        \begin{center}
                \includegraphics[width=11.5cm]{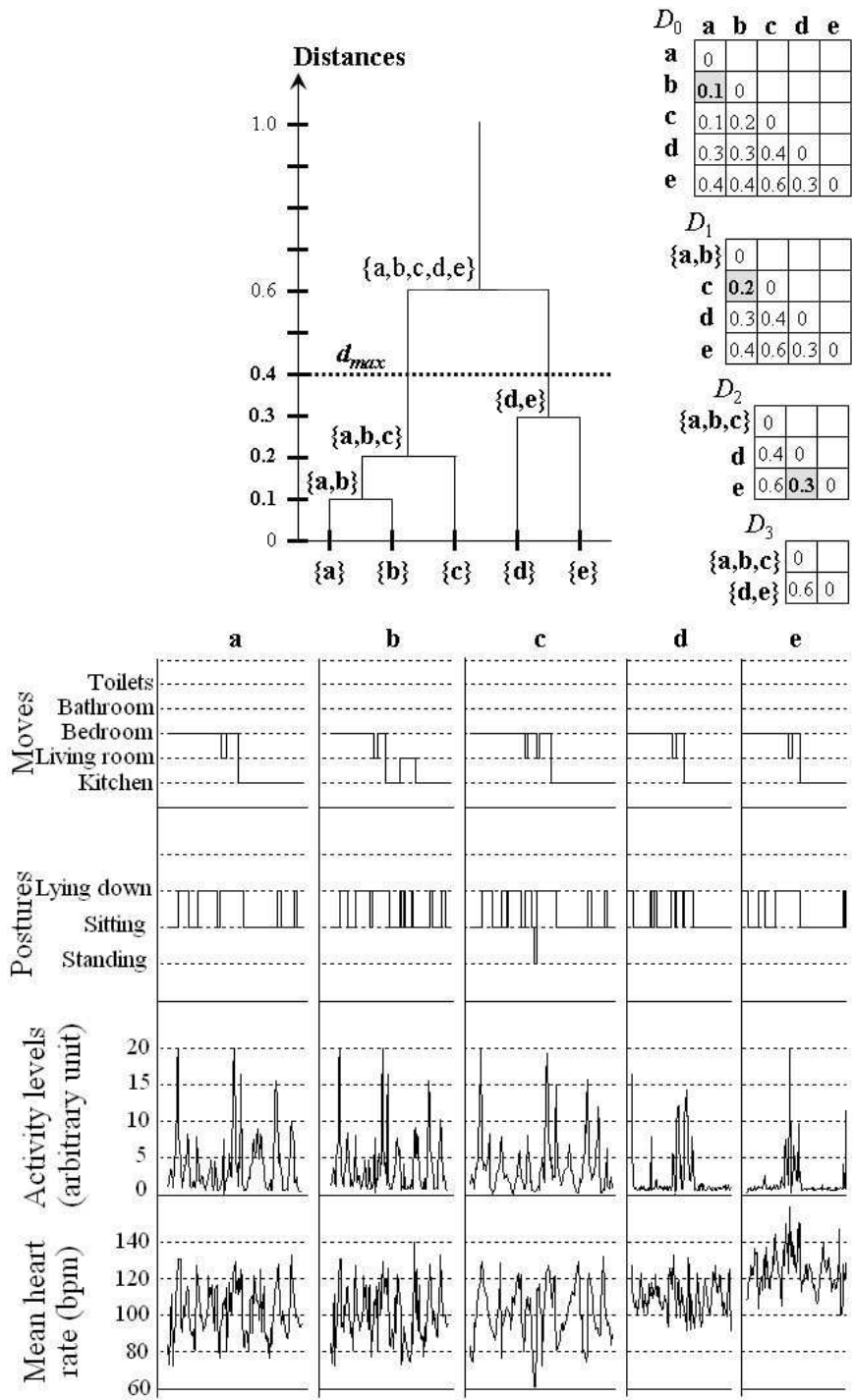}
        \end{center}
        \caption{\small \textbf{Ascending Hierarchical Classification.}}\label{fig_class}
\small The tentative motifs numbered from (a) to (e) are classified into two classes (\{a,b,c\} and \{d,e\}) according to the initial distance matrix $D_0$ and the distance threshold $d_{max}=0.4$.
The corresponding subsequences are displayed on the bottom graphs. They have been generated by a simulation process in the context of home health telecare, including four parameters : (1) the moves of a monitored person, (2) their postures, (3) the activity level, and (4) the mean heart rate.
\end{figure}

The last step is the clustering of tentative motifs into classes representative of any typical ``behavior''. Since the last step, we need a classification method based on an accurate distance measure, that is an actual distance between the original, preprocessed, subsequences corresponding to the identified tentative motifs. We then use the similarity measure defined in section \ref{similarity} for heterogeneous multivariate time-series. Our purpose is to cluster subsequences in groups whose elements are close to all the other ones belonging to the same group, that means the distances are less than a given \emph{distance threshold}. 
Then, we use a \textbf{hierarchical ascending classification} from the distance table between all the tentative motifs, as illustrated on figure \ref{fig_class}. This is an agglomerative technique which starts with single member clusters -- the \emph{tentative motifs} -- successively gathered into classes according to a distance threshold. For the sake of homogeneity and robustness, we use the same distance threshold than when examining the collision matrix. The distance between two classes is defined as the maximum distance observed between all possible pairs of subsequences, one from each class. This ensures we never gather classes containing subsequences whose distance overpass the distance threshold. An additional constraint on the size of any class of \emph{tentative motif} is also added when clustering tentative motifs to reinforce the relevance of extracted patterns.

\begin{figure}[t]
        \begin{center}
                \includegraphics[width=\textwidth]{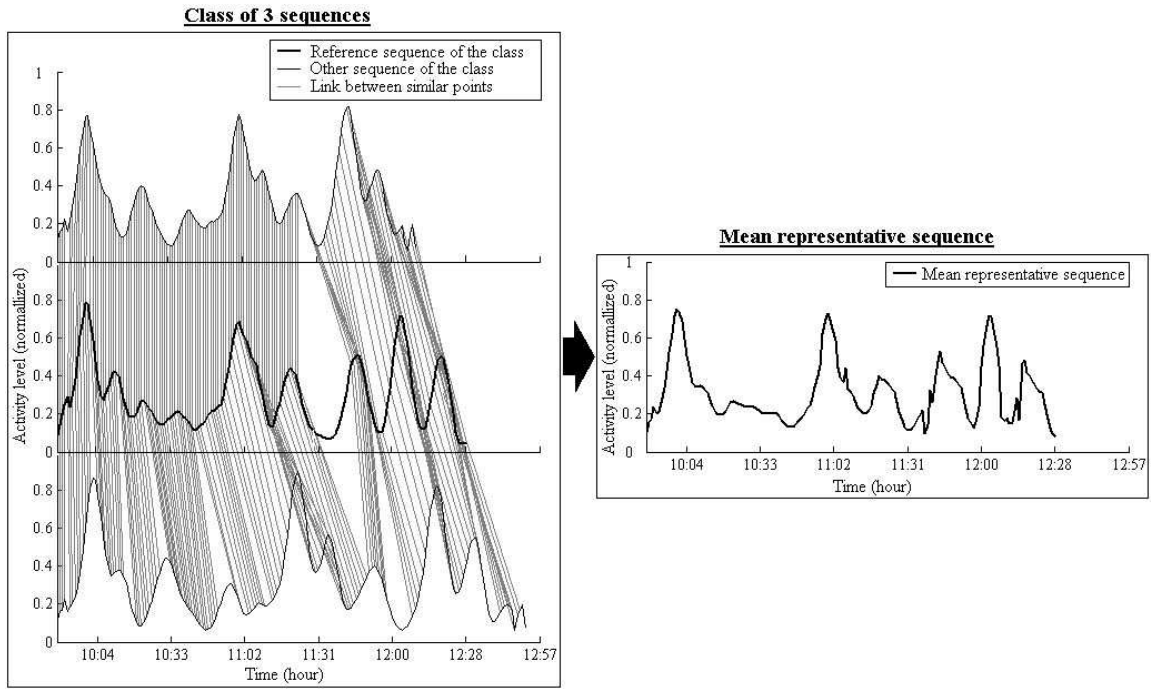}
        \end{center}
        \caption{\small \textbf{Computation of the mean representative sequence of a class.}}\label{fig_represent}
\small The figure above shows how to get the representant of the three one-dimensional sequences of a given class. The computation is performed from the reference sequence, that is the one whose length is closest to the mean sequences length. Each of its points is meant with similar points from the other sequences of the class.
\end{figure}

Once the tentative motifs are clustered into meaningful groups, a mean representative subsequence corresponding to each class -- that is, \emph{motif} -- is computed, as illustrated in figure \ref{fig_represent}. This representative sequence is based on the so called \emph{reference subsequence} whose duration is the closest to the mean duration observed considering all the subsequences of a group. Since based on \emph{LCSS} (see \S \ref{seqdist}), computing a similarity measure between any subsequence of a group and the reference one provides sets of similar vectors associated to each vector of the reference subsequence. The representative subsequence is then defined of same length than the reference sequence, replacing each vector by its mean vector considering the set of associated similar ones.

\section{Experimental Results}\label{experimental_results}

The approach proposed for extracting multidimensional and heterogeneous patterns is experimented in the context of home health telecare. In that section, we first define the experimental context, appropriate to the experimentation of pattern extraction. Testing our approach also requires to define an experimental process, such as relevant measures for evaluating the system's performances. At last, we can discuss the quality of the method and results.

\subsection{Experimental context: home health telecare}\label{experimental_context}

In the purpose of monitoring a person at home, the aim is to learn the person's lifestyle in order to build a sort of behavioral profile, which is sensitive to any critical deviation.
The monitoring system is based on a set of data, recorded at home and in real-time, that may be collected from different classes of sensors: (1) activity (location, position, motion, fall, etc.), (2) environment (temperature, use of doors, window, lighting, etc.), and (3) physiology (blood pressures, weight, etc.). In the definition of these observable parameters, a compromise needs to be found between (a) being easily observable and non invasive, by focusing on the observation of a small set of parameters, and (b) gaining a full appreciation of the person's condition, sensitive to any change in the health status. 
A deterioration of a person's health status usually entails behavioral disorders whose observable symptoms range from an increase in the risk of falls, slowness in executing simple actions, forgetfulness in daily activities, to a global decrease in the person's ability to perform activities of daily living (ADL). Clinical practice has already widely exploited this correlation by estimating a patient's health status in terms of their ability to perform ADL such as getting washed, dressing, or feeding themselves. The usefulness of monitoring some parameters related to the activity of a person is often underlined as being an essential part of any health evaluation \cite{cameron,ramos}, and several projects in home health telecare \cite{peeters,washburn,williams_bb} have already integrated in their concept the assessment of the ADL. Representative of both the activity and the health status, the heart rate is another important and easily observable physiological measure \cite{monod}.

Thus, we decided to consider in a first step of experimentation four parameters that can be defined from a provision of sensors and that are representative of both the heart rate and activity of a person at home, as follows: 
\begin{itemize}
        \item \textbf{Moves:} qualitative, unordered parameter, defining the room occupied by the person at any time. The moves of the person are recorded through infrared motion sensors installed in each room. 
        \item \textbf{Postures:} qualitative parameter, ordered according to the effort required by the posture (``lying down'', ``sitting'', and ``standing''). The postures are inferred from data provided by a set of accelerometers worn by the person.
        \item \textbf{Activity level:} quantitative parameter, in an arbitrary unit. The activity level is measured by a portable accelerometer worn on the chest and estimated through the body acceleration along the anterior-posterior axis \cite{charbonnier}.
        \item \textbf{Mean heart rate:} quantitative parameter, in beat per minute. The mean heart rate is computed from the data recorded by an ECG portable recorder.
\end{itemize}
Thus, we consider a set of heterogeneous parameters for mining meaningful patterns representative of a person's usual behaviors. 

\subsection{Experimental process}\label{experimental_process}

Setting up an experimental process requires to properly define (1) \textbf{which data sets} are appropriate to the context and purpose of the decision-making system, (2) \textbf{which method} are required to build a full experimentation, and (3) \textbf{which performance measures} are the most relevant to objectively evaluate the robustness and efficiency of the system.

\subsubsection{Experimental data sets}\label{experimental_sets}

\subsubsection*{Collecting experimental sets from a simulation process}
The study of any decision-making process requires realistic and accurate data collection. Research projects about home health telecare are as yet only at their first stages of development, and collection of data in realistic environments has just started. Moreover, a full study entails consideration of several profiles of people facing many types of situations. Then, collecting complete and representative sets of data may be a quite hard task, especially to hold data corresponding to rare events. For these reasons, many researchers have turned to simulation as a way to overcome the difficulty of collecting large sets of full experimental data sets. In relation to experimentation, setting up a simulation process enables researchers to have a full and tightly controlled universe of data set, by varying the simulation parameters.

\begin{figure}[t]
\begin{center}
\includegraphics[width=4.5in]{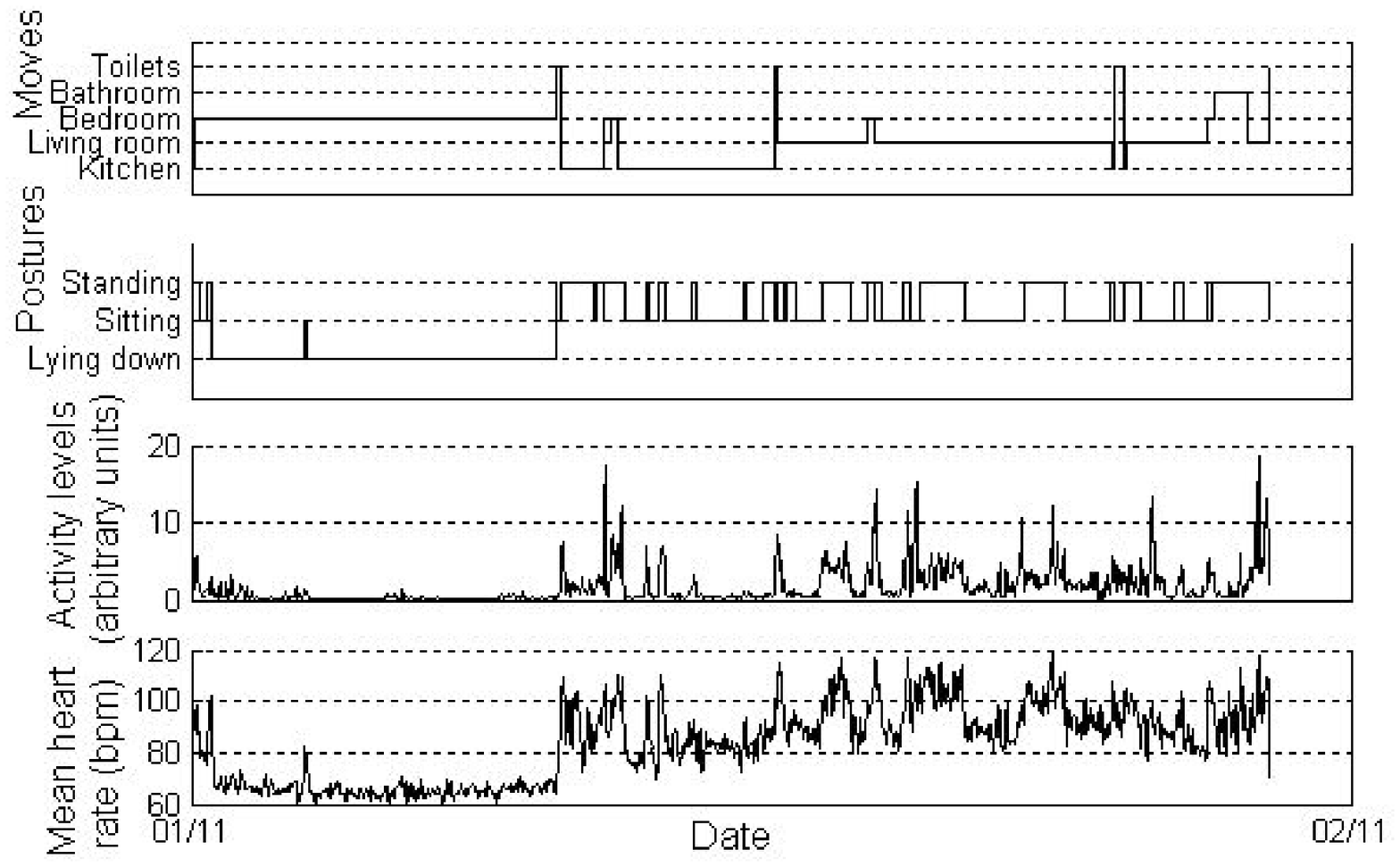}
\end{center}
\caption{\small \textbf{Sequences produced by the simulation process over one day,}}\label{fig_simulation}
\small including four parameters that are, from top to bottom: (1) moves, (2) postures, (3) activity levels, (4) mean heart rate. Data are available every minute.
\end{figure}

For these reasons, we have set up a simulation process for generating realistic sequences corresponding to the experimental parameters \cite{duchene_bb}: moves, postures, activity level, and mean heart rate. 
The simulation process is designed to preserve the problem's complexity, that is especially the joint variations of the parameters. The sequences produced by simulation must also be representative of a person's habits at home, that means they include every day subsequences corresponding to the presence of basic activities of daily living like sleeping, eating three times a day, getting washed. Considering these requirements and relative influences of simulated parameters, the simulation model is defined using a cascade structure, and run in four steps to successively generate time-series corresponding to: (1) the moves of the subject in a given period of time, (2) their successive postures, (3) the sequences of the activity levels, and (4) the values of the mean heart rate \cite{duchene_bb}. A sample of data produced by the simulation process over one day is shown on figure \ref{fig_simulation}.
Our objective is then to identify from these sequences of data ``high level patterns'' corresponding to usual behaviors of the person at home, especially under the presence of noise.

\subsubsection*{Defining appropriate sequences for experimenting pattern extraction}
In order to validate our approach to pattern extraction, we need to objectively check the relevance of the motifs extracted from time-series. That requires to know \textit{a priori} which motifs are represented within the time-series and the location of their instances, so that we can evaluate the performances of the system. The aim is then to introduce instances of predefined motifs in sequences containing at first no pattern. Due to the simulation method, the way of getting time-series containing no significant patterns while remaining realistic in terms of the joint variations of the parameters is described as follows:
\begin{itemize}
        \item First, generation of random moves, so that data include no pattern representative of the person's activities of daily living.
        \item Second, simulation of relevant sequences of values for the other parameters according to these random moves, so that the multidimensional sequences remain realistic anyway.
\end{itemize}
We also need to identify realistic motifs considering the monitoring of a person at home. This time, the simulation process is used to generate sequences corresponding to living habits, in which we can randomly select subsequences that can be interpreted in terms of the person performing a typical activity for a given time. The length of subsequences is randomly selected in meaningful bounds (from 30 minutes to 2 hours for instance). We also add a constraint on the selected subsequences to ensure they can be interpreted as the person performing a given activity, and not only one or two elementary action. Selected subsequences must then be represented -- using the representation step defined for motifs extraction (see \S \ref{representation}) -- by a sequence of at least 4 symbols. 

The last step is then to create and introduce some instances of these predefined motifs in the non-pattern sequences previously generated.
The introduction of motifs' instances is guided by specific characteristics of patterns identified in our experimental context (see \S \ref{guidelines}). We especially need to take account of possible imprecise matches, outliers, translation and stretching in time between instances of a same motif. Therefore, we define several types of noise that can be introduced in a subsequence representing a motif to get instances of the same motif, as follows:
\begin{itemize}
        \item \textbf{Noisy values.} Given that we consider sequences representative of human behaviors, instances of a same motif fit only in their global trends. Then, we may define a high rate of noise in the values.
        \item \textbf{Interruptions.} Any main activity of daily living may be interrupted by a secondary task like answering to the phone or going to the toilets. As much as this additional task remains of short duration, we would like to recognize the main task anyway. That requires to experiment the algorithm using instances of motifs including short sequences of \emph{outliers}, representing disruptions in the global trend of variation.
        \item \textbf{Stretching in time.} Dealing with human behaviors, a same activity does not always last the same duration. That is the reason why we may introduce instances of different lengths corresponding to a same motif.
        \item \textbf{Translation in time.} Even usual at home, an activity is not always performed exactly at the same time. That is the reason why we introduce instances of a same motif anywhere in the non-pattern time-series.
\end{itemize}

\subsubsection{Experimental method}

Setting up an experimental process is relevant for evaluating the system's performance at two levels: 
\begin{enumerate}
        \item \textbf{Quality of the method:} that includes to validate each stage of the proposed approach and to define appropriate values for the parameters required. We are especially interested in defining the sensitivity of each parameter, and the way relevant values can be identify and validated.
        \item \textbf{Quality of the results:} once the system is properly set up, the aim is to evaluate the performances of the system especially under the strong presence of noise. We aim at study the influence on introducing highly noisy instances of given motifs, considering all possible noise: variability in values, outliers, stretching and translation in time.
\end{enumerate}

\subsubsection{Performance measures}\label{perf_measures}

Setting up an experimental process includes to define appropriate performance measures. In our context, the objective evaluation of the robustness and efficiency of our approach to motifs extraction is performed at two levels: (1) Identification of frequent subsequences within time-series containing both pattern and non-pattern signals, that is the \emph{performance of tentative motifs extraction}; and (2) Classification of these subsequences into motifs, that is the \emph{performance of clustering tentative motifs}.
We then define means of evaluating the sensibility and specificity for these two stages of motifs extraction.

\subsubsection*{Tentative motifs extraction}

Defining a measure of sensibility and specificity for this stage aims at evaluating the ability of this algorithm to (a) well identify as tentative motifs subsequences corresponding to instances of motifs (sensibility), and (b) not define as tentative motifs subsequences corresponding actually to non-pattern signals (specificity). Sensibility ($Se$) and specificity ($Sp$) are computed from rates of true/false positive/negative (labeled $TP$, $FP$, $TN$, $FN$), considering one of the two following complementary hypothesis for each point of the original sequence: ``the point belong to an instance of a motif'' and ``the point does not belong to an instance of a motif''.
        \[Se = \frac{TP}{TP+FN} \mbox{ and } 
          Sp = \frac{TN}{TN+FP}
\]

\subsubsection*{Clustering into motifs}

At this stage, the aim is to evaluate the ability of the algorithm to properly cluster the tentative motifs into motifs, that is (a) to gather all the instances of a same motif in one class (sensibility) and (b) to gather in one class only some instances of a same motif (specificity). Sensibility and specificity of classification algorithms of $N$ vectors are determined using confusion matrix. A confusion matrix $C$ of dimension $m \times n$ represents the results of an algorithm that clusters $N$ vectors, corresponding \textit{a priori} to $m$ motifs, in $n$ classes. Each value $c_{ij}$ represents to the number of vectors belonging to class $j$ that actually correspond to instances of motif $i$. Considering a confusion matrix, the sum over any row $i$ is the number of instances of motif $i$, and the sum over any column $j$ is the number of vectors in class $j$.

Due to the way of extracting the tentative motifs, our context presents some specific features in comparison to defining ``simple'' performance measures of clustering algorithms.
\begin{itemize}
        \item The ``right'' elements to be clustered -- that is all and only subsequences corresponding effectively to motifs -- may not be available, depending on the performance of tentative motifs extraction. As a consequence, the sum over each row $i$ may be lower than the number of instances of motif $i$, and the sum over each column $j$ of the confusion matrix lower than the number of elements in class $j$. Moreover, an instance of a motif may be recognize as more than one tentative motif, so that these sums may also be greater than the usual values expected.
        \item Because we use unsupervised classification, the number of classes as output of the clustering algorithm might be different from the effective number of motifs, that is $m \neq n$. Analysis of clustering performances reported in the literature does not however usually consider that issue.
\end{itemize}
Considering these assumptions entails the definition of sensibility and specificity as follows:
\begin{itemize}
        \item \textbf{Sensibility:} \emph{``All the instances of a motif must be gathered in one class, as unsegmented subsequences -- that is, an instance of a motif is associated to only one subsequence of the class.''}
        \item \textbf{Specificity:} \emph{``All the elements of a class must be representative of a same motif, as unsegmented subsequences -- that is, only one element of a class is associated to each instance of the corresponding motif.''}
\end{itemize}
Missing some motifs instances, recognizing some instances as several subsequences, or failing in properly clustering the tentative motifs decrease these performance measures.
The proposed definitions of sensibility and specificity are implemented using the concept of \emph{entropy}. A null value of entropy represents the perfect order, which should correspond to a maximum value, 1, of sensibility and specificity: tentative motifs extraction and clustering are perfectly done. Sensibility is related to the well identification and clustering of motifs instances, so that it is defined as a measure of entropy over each row $i$ of the confusion matrix. Specificity is related to the homogeneous composition of each class, so that it is defined as a measure of entropy over each column $j$ of the confusion matrix. These indexes could then be roughly defined as:
\begin{equation}\label{se_sp_base}
Se_i = 1 + \frac{1}{log(n)} \cdot \sum^{n}_{j=1}\frac{c_{ij}}{m_i}\cdot log\left(\frac{c_{ij}}{m_i}\right)
\mbox{ and }  
Sp_j = 1 + \frac{1}{log(m)} \cdot \sum^{m}_{i=1}\frac{c_{ij}}{n_j}\cdot log\left(\frac{c_{ij}}{n_j}\right)
\end{equation}
\begin{center}
\begin{tabular}{rlll}
                        where & $m$ &is the&number of motifs expected\\
                        & $n$ &&number of classes extracted\\
                        & $m_i$ &&number of instances of motif $i$\\
                        & $n_j$ &&number of elements in class $j$.\\
\end{tabular}
\end{center}
The values $\frac{1}{log(n)}$ and $\frac{1}{log(m)}$ correspond to the maximum entropy of a system of $n$ and $m$ states respectively, so that $Se_i$ and $Sp_j$ values are into [0,1]. 

\noindent Using equations (\ref{se_sp_base}), the following properties must be verified:
\begin{equation}\nonumber
\sum^{n}_{j=1}\frac{c_{ij}}{m_i} = 1 
\mbox{ and } 
\sum^{m}_{i=1}\frac{c_{ij}}{n_j} = 1
\end{equation}
However, in our context, we can miss some instances of motifs, or recognize them as divided into several subsequences. Consequently, the values of $\sum^{n}_{j=1}{c_{ij}}/{m_i}$ and $\sum^{n}_{j=1}{c_{ij}}/{n_j}$ might be either lower -- when missing some instances -- or greater -- when tentative motifs split up -- than 1, and we must adapt the formulas (\ref{se_sp_base}).

In order to deal with missing instances, we introduce a notion of \emph{``recognition rate''} for each motif $i$, $\rho e_{i}$, and each extracted class $j$, $\rho p_{j}$ defined as:
\begin{equation}\label{eq_rho}
           \rho e_{i} = \frac{\sum^{n}_{j=1}c_{ij}}{m_i} \mbox{ and }
                 \rho p_{j} = \frac{\sum^{m}_{i=1}c_{ij}}{n_j}
\end{equation}
These parameters are used as weights in computing sensibility and specificity, so that the values of sensitivity and specificity decrease in case some instances of a motif are not identified, or some elements of a class are not representative of a motif's instance. Equations (\ref{se_sp_base}) then become:
\begin{eqnarray}\label{se_sp_adapt1}
Se_i = \rho e_{i}\cdot\left(1 + \frac{1}{log(n)} \sum^{n}_{j=1}\frac{c_{ij}}{\sum^{n}_{j=1}{c_{ij}}}\cdot log\left(\frac{c_{ij}}{\sum^{n}_{j=1}{c_{ij}}}\right)\right), \\
Sp_j = \rho p_{j}\cdot\left(1 + \frac{1}{log(m)} \sum^{m}_{i=1}\frac{c_{ij}}{\sum^{m}_{i=1}{c_{ij}}}\cdot log\left(\frac{c_{ij}}{\sum^{m}_{i=1}{c_{ij}}}\right)\right).\label{se_sp_adapt2}
\end{eqnarray}

Furthermore, to get an appropriate measure to deal with motifs' instances possibly discovered as several tentative motifs, we introduce the notion of \emph{``split rate''} for each instance $k$ of motif $i$, $1/\eta_{ik}$, where $\eta_{ik}$ is the number of tentative motifs associated to instance $k$ of motif $i$. When filling in the new confusion matrix $C^{'}$, we then do not consider adding 1 to $c_{ij}$ when instance $k$ of motif $i$ is recognized in class $j$, but adding $1/\eta_{ik}$ to $c^{'}_{ij}$. Consequently, $\sum^{n}_{j=1}c^{'}_{ij}$ represents the number of instances of motif $i$ that are well identified, even if not well classified, and we have $\sum^{n}_{j=1}c^{'}_{ij} \leq m_i$ ; and $\sum^{m}_{i=1}c^{'}_{ij}$ represents the number of distinct motifs instances represented in class $j$. This last value is not necessarily an integer because an element of a class can represent only one part of a motif's instance, the other part(s) being classified in other classes.

\begin{figure}[t]
\begin{center}
\includegraphics[width=12cm]{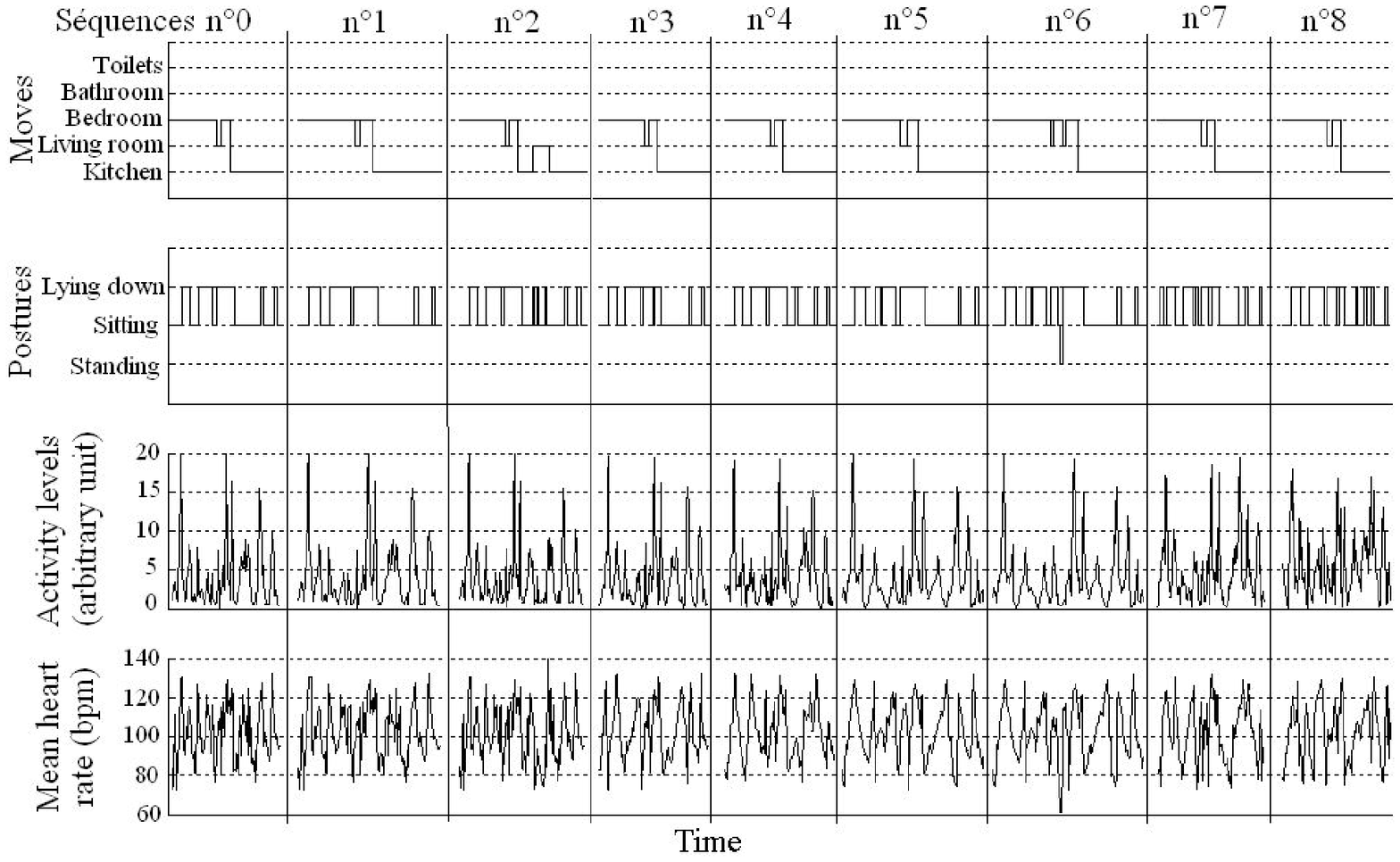} 
\includegraphics[width=14.5cm]{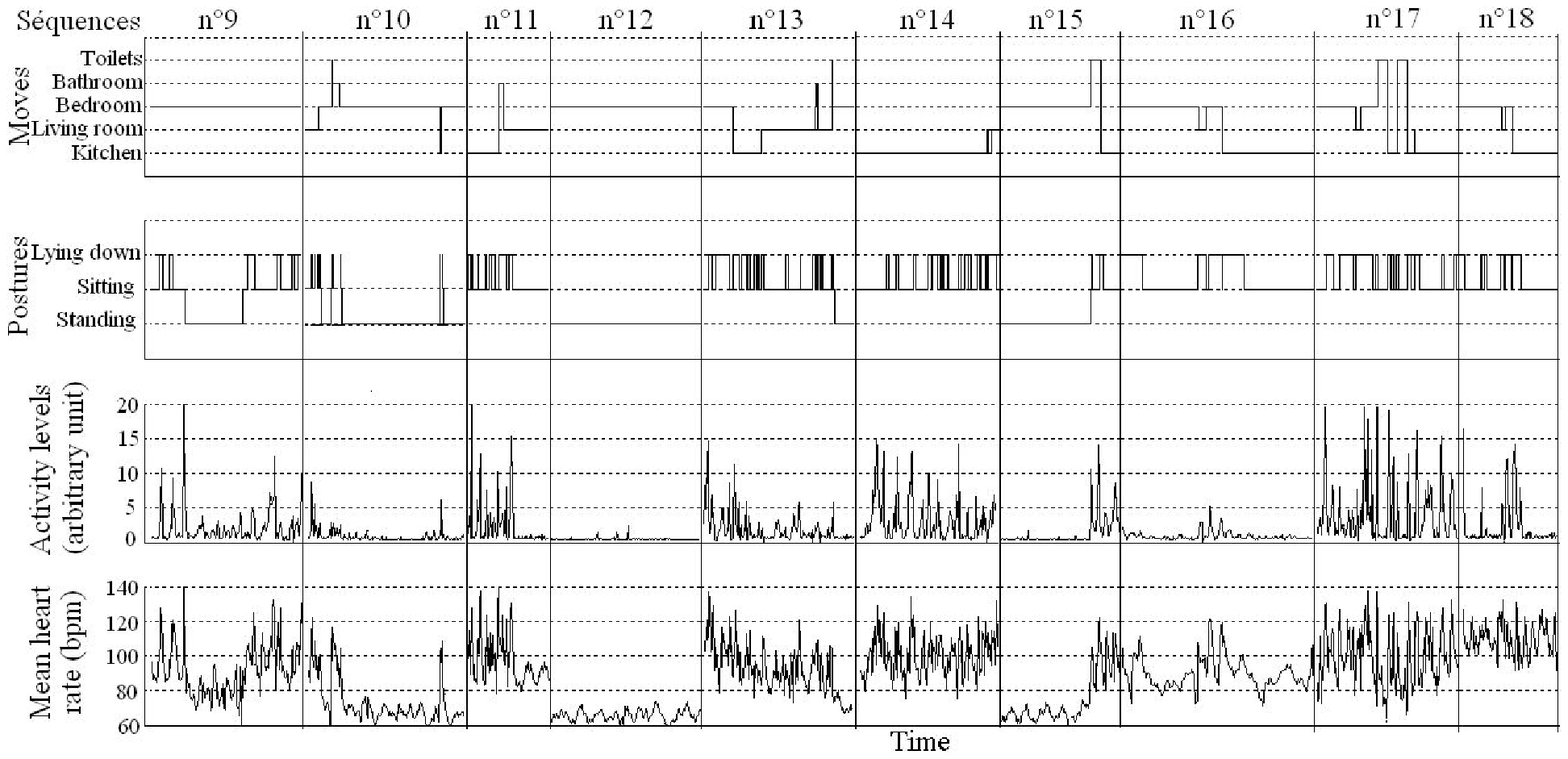} 
\end{center}
\caption{\small \textbf{Experimental sequences.}}\label{fig_seq_class}
\small From top to bottom, sequences 1 to 8 should be considered close to sequence 0 (class 0), and sequences 9 to 18 far from this reference sequence (class 1).
\end{figure}

Sensibility and specificity are then defined using equations similar to (\ref{se_sp_adapt1}) and (\ref{se_sp_adapt2}), replacing $c_{ij}$ by $c^{'}_{ij}$. We also re-defined the \emph{``recognition rates''} from (\ref{eq_rho}) to take into account the new definition of the confusion matrix, $C^{'}$, as:
        \[\rho e_{i} = \frac{\sum^{n}_{j=1}c^{'}_{ij}}{\sum^{n}_{j=1}c^{'}_{ij} + m^{'}_{i}} 
        \mbox{ and }
                \rho p_{j} = \frac{\sum^{m}_{i=1}c^{'}_{ij}}{\sum^{m}_{i=1}c^{'}_{ij} + n^{'}_j},\]
\begin{center}
\begin{tabular}{rlll}
                        where & $m^{'}_i$&is the&number of instances of motif $i$ not identified, in any class\\
                                          & $n^{'}_j$ &&number of elements in class $j$ representative of any motif.\\
\end{tabular}
\end{center}

\noindent We also introduce an index of segmentation of motifs recognition, $\lambda$, according to equation \ref{eq_lambda}. A maximum value of 1 means that the whole motifs instances are well recognized.
\begin{equation}\label{eq_lambda}
        \lambda = \frac{1}{n\cdot m}\sum^{m}_{i=1} \sum^{n}_{j=1} \frac{c^{'}_{ij}}{c_{ij}}
\end{equation}

\noindent We can at last define mean values of sensibility and specificity, as:
\[Se = \frac{1}{m}\sum^{m}_{i=1}Se_i
        \mbox{ and }
        Sp = \frac{1}{n}\sum^{n}_{j=1}Sp_j.
\]

\begin{figure}[t]
\begin{center}
\includegraphics[width=8.4cm]{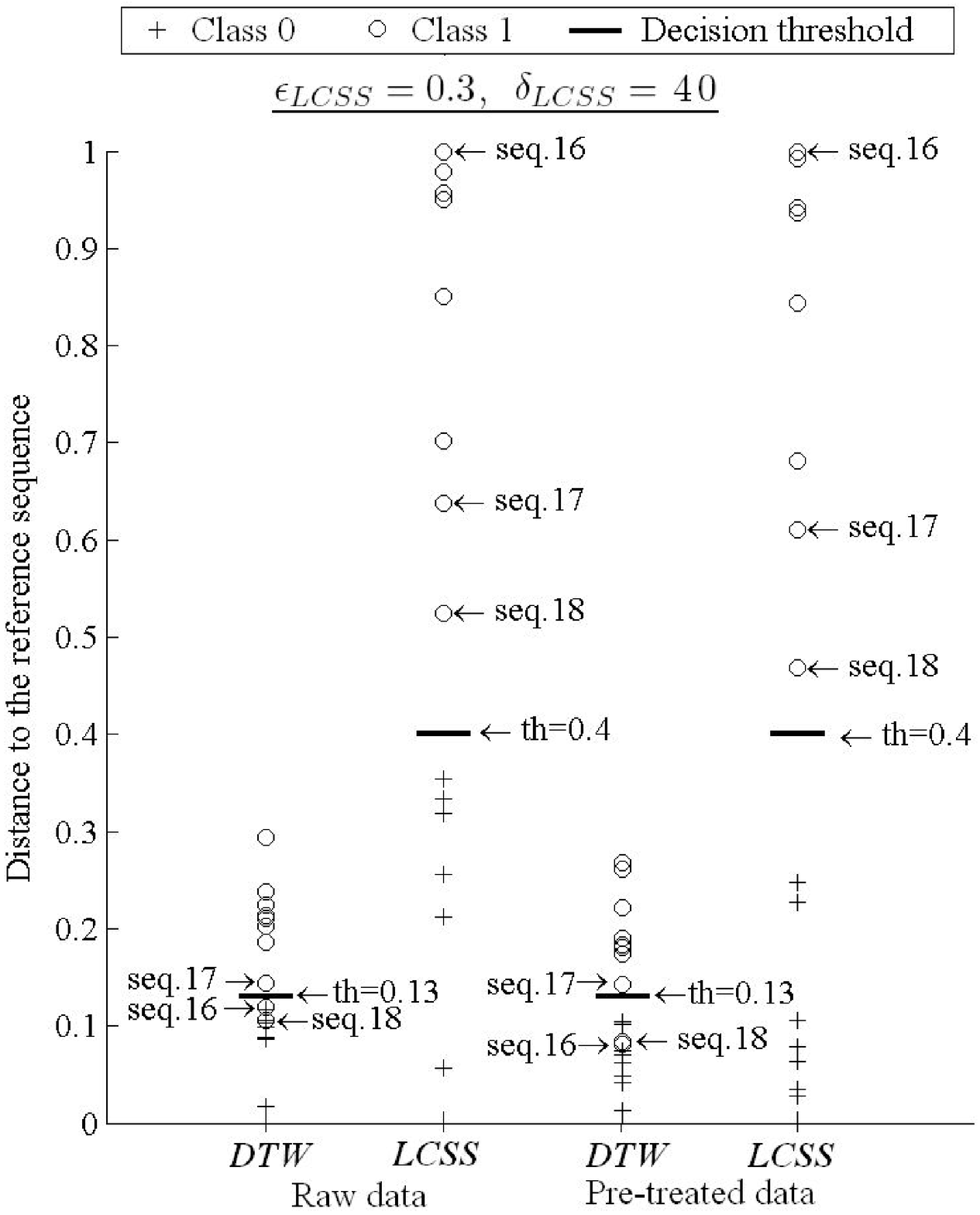}
\end{center}
\caption{\small \textbf{Distances between sequence 0 and the other experimental sequences,}}\label{fig_cmp_LCSS_DTW}
\small from either raw or pre-processed data, and using \emph{DTW} and \emph{LCSS} distances. Classes 0 and 1 correspond to the expected classification.
\end{figure}

\begin{figure}[h]
\begin{center}
\includegraphics[width=12.5cm]{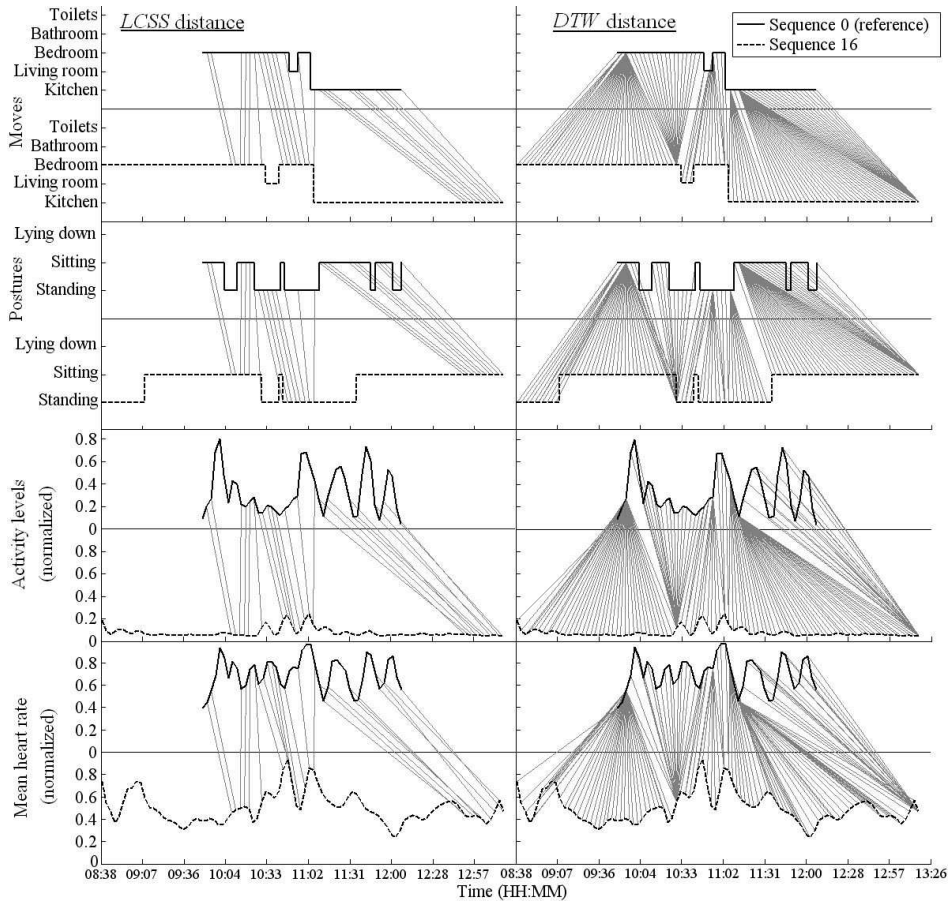} 
\end{center}
\caption{\small \textbf{Pairs of points considered as similar when computing \emph{LCSS} and \emph{DTW} distances.}}\label{fig_cmp_LCSS_DTW_seq16}
\end{figure}

\begin{figure}[t]
\begin{center}
\includegraphics[width=\textwidth]{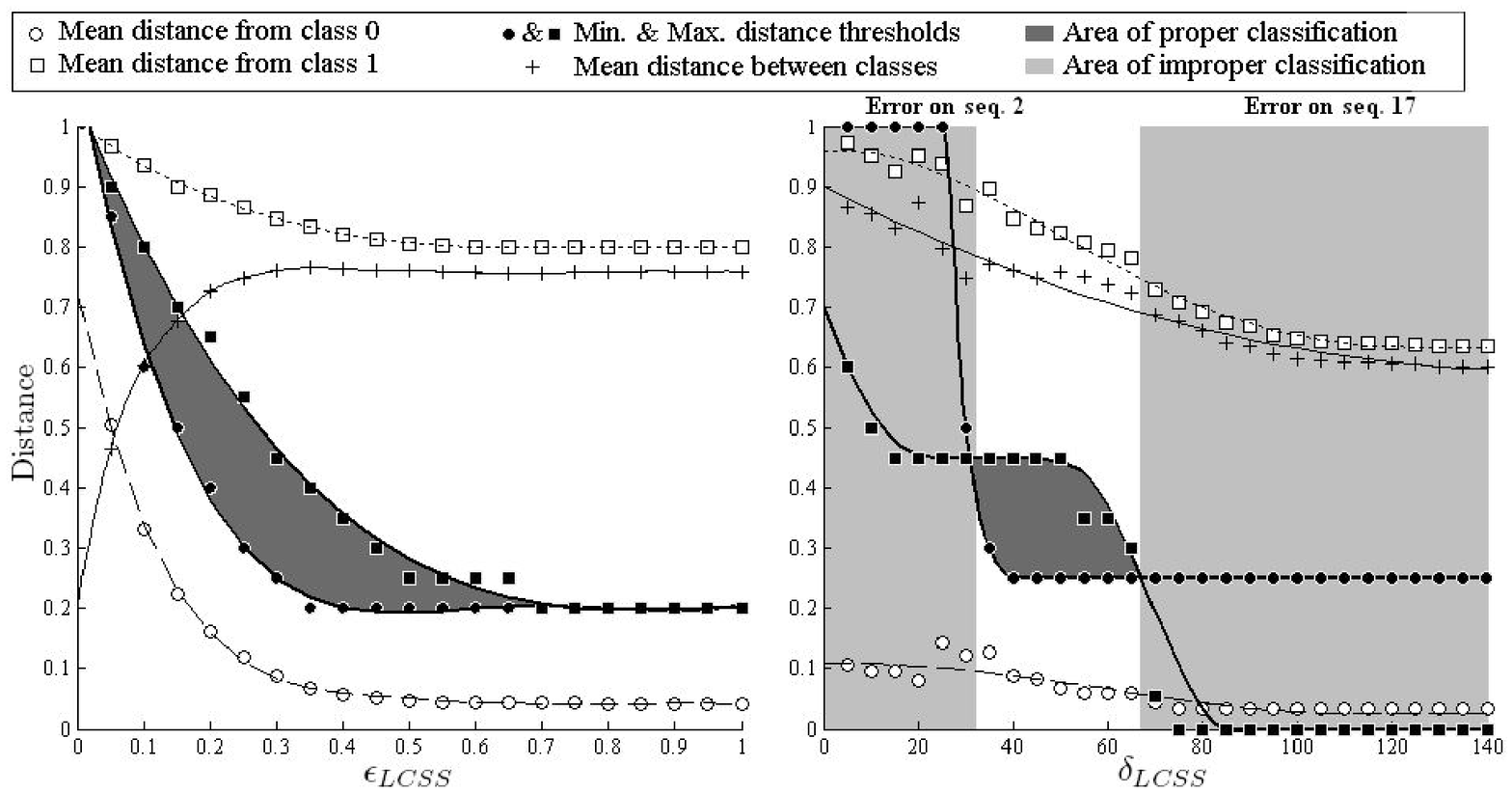}
\end{center}
\caption{\small \textbf{Distances observed according to selected values of $\epsilon_{LCSS}$.}}\label{fig_dist_delta_epsilon}
\small These results are obtained when classifying sequences 1 to 18 in class 0 or 1 depending on whether or not close to reference sequence 0 according to a given distance threshold. The graph presents the mean distances observed for class 0 and 1 between each sequence of the class and the reference sequence 0. The minimum and maximum decision thresholds required for a proper classification of sequences in class 0 or 1 is also plotted, such as the mean distance observed between classes. In these samples $\delta_{LCSS}$ has been selected as not restrictive.
\begin{itemize}
        \item The left graph shows the variation of these distances when the similarity threshold on values -- $\epsilon_{LCSS}$ -- is set between 0 to 1, and $\delta_{LCSS}=40$ minutes.
        \item The right graph shows these variations for increasing values of the similarity threshold in time -- $\delta_{LCSS}$ (in minutes) -- and $\epsilon_{LCSS}=0.3$.
\end{itemize}
\end{figure}

\subsection{Quality of the method}\label{method_quality}

Given the high complexity of the problem and the need for several successive steps of analysis, we integrate several levels of validating the proposed approach, that is in consideration to the following steps: (1) Defining a similarity measure, (2) Representation, and (3) Mining operations. Validation is performed from both experts and mathematical and statistical analysis, depending of whether or not appropriate data are available for objective comparisons.

\subsubsection{Similarity measure}\label{valid_similarity}

The approach defined for computing the distance between multivariate and heterogeneous time-series is especially experimented and validated under the strong presence of noise. We compare the performance of our approach, which involves \emph{LCSS}, to the use of \emph{DTW} distances. 
The distance between time-series is expected to generate low values between sequences corresponding to the realization of a same activity in same conditions, and higher values otherwise. Two experimental sets consist of (see figure \ref{fig_seq_class}):
\newcounter{foo} \setcounter{foo}{1} 
\begin{description}
 \item [(\arabic{foo})] \textbf{Sequences 0 to 8 representative of a given activity} --- getting ready in the morning, generated from a reference sequence (sequence 0) by adding noise of three types: stretching in time, variability in values, interruptions (consecutive outliers).
 \addtocounter{foo}{1}
 \item [(\arabic{foo})] \textbf{Sequences 9 to 18 representative of other activities} like sleeping, having a meal, having a quiet activity, including three sequences (sequences 16 to 18) corresponding to the reference activity (same moves) but carried out in bad conditions, that is (16) slowness, (17) long worrying interruptions, and (18) high values of the mean heart rate. These abnormal behaviors may be detected if sequences 16 to 18 are not considered as representative of sequence 0.
\end{description}
The experimental process aims at classifying these sequences using a threshold on the distance to sequence 0. An appropriate distance may be able to properly discriminate the sequences: 1 to 8 associated to class 0, and 9 to 18 to class 1. We use both \emph{DTW} and \emph{LCSS} distances for comparison (see \cite{keogh} for a clear review of \emph{DTW} principle), and in each case the distances are computed from both raw and pre-processed data --- that is sampling rate reduction to speed-up the computation, and filtering to remove some noise. Preliminary experimentations were required to define relevant values for the \emph{LCSS} parameters ($\epsilon$,$\delta$) in the context of our application. 

The classification results are presented on figure \ref{fig_cmp_LCSS_DTW}. As a general comment, we notice that \emph{DTW} distances are really lower than \emph{LCSS} ones, due (1) to different orders of computation --- 1 for \emph{LCSS} and 2 for \emph{DTW}, and (2) to possible multiple associations of any point using \emph{DTW}, so that the distance may remain quite low.

The superiority of \emph{LCSS} over \emph{DTW} is pointed out by the results matching the expected classification only in the case of using \emph{LCSS}. Using \emph{DTW} distance fails in properly classifying critical sequences 16 and 18. The behavior of both \emph{LCSS} --- $\delta$ set with no restriction in time for associating points, as it is using \emph{DTW} --- and \emph{DTW} when comparing sequences 0 and 16 is illustrated on figure \ref{fig_cmp_LCSS_DTW_seq16}. \emph{DTW} allows for multiple associations, and all points must be matched, based on a minimum distance criterion. Then, because the sequences of moves and postures are very close, the poor number of points corresponding to low activity levels and mean heart rate in sequence 0 are associated to the large number of such points in sequence 16, and reciprocally for high values of activity levels and mean heart rate. That results in a low number of pairs corresponding to large distances, so that the distance between the two sequences remains low. The strength of \emph{LCSS} is to base the similarity of points on a threshold criterion, allowing outliers, and excluding overlapping pairs. A higher \emph{LCSS} distance is even obtained for sequence 16 by restricting the value of $\delta$.

We also notice that the two classes are better separated when computing the distances from the preprocessed data. Filtering the sequences indeed results in removing at least part of the variability in the values.\\
\noindent Then, the key parameters in defining the similarity measure are as follows:
\begin{itemize}
        \item \textbf{Maximum temporal difference ($\delta_{LCSS}$)} between two points so that they can be considered as similar. 
        \item \textbf{Maximum difference in the values ($\epsilon_{LCSS}$)} of two points so that they can be considered as similar.
\end{itemize}
Defining $\delta_{LCSS}$ and $\epsilon_{LCSS}$ restrains the selection of a relevant decision threshold to properly classify sequences, as illustrated on figure \ref{fig_dist_delta_epsilon}. For instance, increasing values of $\epsilon_{LCSS}$ result in decreasing distance measures, so that an appropriate decision threshold can be defined lower and lower. There is however a stabilization of values as $\epsilon_{LCSS}$ ranges over 0.5. Decreasing the value of $\delta_{LCSS}$ also results in increasing values of distances.

\subsubsection{Representation}

The stage of representation is validated at two levels:
\begin{enumerate}
        \item Checking with experts that the representation step preserves within the sequences the trends of variation they consider as important in identifying the behavioral profile of a person.
        \item Analyzing the influence of each step of representation in preserving these fundamental trends according to the purpose of study. 
\end{enumerate}
The step of representing temporal sequences includes preprocessing, discretization and aggregation. Validation is performed using intuitive knowledge from experts. The representation steps are anyway highly restrained by the context and purpose of the decision issue, so that there are not many choices in selecting appropriate values for the parameters involved at this stage.
Considering several experimentations with the few possible values of parameters, the representation of raw time-series (generated by simulation) is intuitively evaluated according to the purpose of preserving the global trends while removing insignificant variations in terms of studying the activities of daily living of a person. We then define the key parameters as follows:
\begin{itemize}
        \item \textbf{Filter type and length:} we use a mean weighted filter, so that highlights the global trends while preserving the peaks in the values.
        \item \textbf{Rate of temporal reduction:} some analysis show that a rough temporal reduction of initial time-series may remove critical points, especially peaks in the values. Anyway, at the end of the representation step, the aggregation of time-series produces approximately the same number of symbols representing the original sequence, whatever a temporal reduction. We then decide to rely on meaningfully aggregating the successive vectors to reduce the sequences length.
        \item \textbf{Number of discretization intervals:} the meaningful number of discretization intervals for values of quantitative parameters may be approximately defined by experts. Considering the activity level and mean heart rate, the intuitive qualification of possible variations as ``resting'', ``low'', ``moderate'', and ``high'' guide us to define four intervals of discretization. This rough idea could be refined considering the system's performances. We use k-means algorithm to determine appropriate bounds of discrete intervals for each monitored person (see \S \ref{discretization}). The obtained results are roughly in agreement with related academic knowledge, as illustrated in table \ref{tab_motifs_discret}.
\vspace{0.1cm}
\begin{table}[h]
        \begin{center}
                \begin{tabular}{|c||l|c||c|c|}
                        \cline{2-5}
                        \multicolumn{1}{c|}{ } & \multicolumn{2}{c||}{Academic knowledge} & \multicolumn{2}{c|}{Experimentation}\\
                        \cline{2-5}
                        \multicolumn{1}{c|}{ } & \textbf{Activity} & \textbf{Mean heart} & \textbf{Activity} & \textbf{Mean heart}\\
                        \multicolumn{1}{c|}{ } & \textbf{level} & \textbf{rate} & \textbf{level} & \textbf{rate}\\
                        \hline
                        1 & \emph{Rest} & $\approx 65$ & $< 1.8$ & $< 75$\\
                        2 & \emph{Very light} & $< 75$ & $1.8$ to $3.8$ & $75$ to $92$\\
                        3 & \emph{Light} & $75$ to $100$ & $3.8$ to $7$ & $92$ to $104$\\
                        4 & \emph{Moderate} & $100$ to $125$ & $> 7$ & $104$ to $120$\\
                        \hline
                \end{tabular}
        \end{center}
\caption{\small \textbf{Discretization intervals for the activity level (arbitrary unit) and the mean heart rate (beat per minute) in comparison with empirical boundaries \cite{monod}.}}\label{tab_motifs_discret}
\end{table}
        \item \textbf{Minimum distance threshold:} we use of minimum distance of zero as a threshold for aggregating successive vectors. That means aggregation in one symbol is allowed along subsequences where there is no significant variations of the parameters, so that we can intuitively assume that the person is performing a same ``action'' all along the corresponding duration. Increasing the minimum distance does not seem appropriate because, especially considering qualitative parameters, a change in a value corresponds to a change in the room occupied or the posture, which represents intuitively a change in the ``elementary action'' performed.
\end{itemize}
As a consequence, the number of discretization intervals is the only parameter whose value could be refine by studying its influence on the system's performances, but the possible values are highly constrained anyway. The relevance of considering 4 discrete intervals in our context has been then confirmed by experimentations. For further experimentations, we then decide to use the most intuitively appropriate values of parameters, that is: mean weighted filter, no temporal reduction, 4 intervals of discretization, and a null minimum distance threshold for aggregation.

\subsubsection{Mining operations}\label{quality_method_mining}

Mining operations include feature mining -- that is, projections, collision matrix examination, and tentative motifs extraction -- and clustering. The parameters involved at that stage are differently constrained in their values and have specific influence on the system's performances.

Due to the experimental context, we have identify three highly constrained parameters, so that we can select \textit{a priori} the most appropriate value. Some experimentations can however be performed to possible refine this intuitive choice.
\begin{itemize}
        \item \textbf{Number of symbols} considered to define the length of basic subsequences defined for projections. This length should correspond to the minimum number of symbols defining a motif. Thus, this parameter highly depends on the level of representation of the original sequence in terms of successive aggregated vectors -- the symbols. In our context, that corresponds to the minimum number of ``actions'' successively performed by a person in their activities of daily living, which we have defined intuitively as being 4 symbols. Selecting a lower value, like 3 symbols, could be appropriate, but, reversely, we might miss some motifs' instances. 
        \item \textbf{Projection mask} defines the number of symbols, as well as the number of parameters of each symbol, not considered when comparing basic subsequences after projection. That means similar subsequences could differ in these numbers of symbols and components. These parameters are consequently determined according to the rate of noise allowed between two similar subsequences. Given that we consider 4 dimensions for defining symbols, and 4 symbols in basic sequences, the projection mask is defined so that we project 3 symbols and 3 components of these symbols. This choice is validated by the intuitive idea of low collisions numbers between different sequences in the context of observing the mean percentage of collisions between sequences of class 0 or class 1 and the reference sequence 0 (see figure \ref{fig_seq_class}), as illustrated on table \ref{tab_dim_proj}.
\vspace{0.1cm}
\begin{table}[h]
\begin{center}
\begin{tabular}{|c||c|c|c||c|c|c|}
\hline
\textbf{Size}&\multicolumn{3}{|c||}{\textbf{Mean \% of collisions}}&\multicolumn{3}{|c|}{\textbf{Median \% of collisions}}\\
\cline{2-7}
\textbf{of projection}&Class 0&Class 1&Gap&Class 0&Class 1&Gap\\
\hline
\hline
\textbf{3} $\times$ \textbf{3} &\textbf{46.1}&\textbf{8.5}&\textbf{37.6} &\textbf{45.5}&\textbf{3.5}&\textbf{42.0} \\
\hline
$2\times 3$&56.2&18.3&37.9&61.5&14.5&47.0\\
\hline
$3\times 2$&58.2&20.3&37.9&64.0&21.0&43.0\\
\hline
$2\times 2$&67.5&34.5&33.0&78.5&39.5&39.0\\
\hline
\end{tabular}
\end{center}
\caption{\small \textbf{Percentage of collisions observed between sequences of class 0 and 1 and the reference sequence 0 according to the size of the projection mask.}}\label{tab_dim_proj}
\small The size of the projection mask correspond to the number of projected symbol $\times$ the number of projected parameters per symbol.
\end{table}
        \item \textbf{Number of projections} performed to build the collision matrix. The number of projections should ensure that (1) we do not get hazardously a high number of collisions (specificity), and (2) similar subsequences as defined in our context correspond to a high number of collisions (sensibility). With an increasing number of symbols defining basic subsequences, dimensionality, and rate of possible noise, then should also increase the number of projections performed to get significant results. Given that these influence factors are well defined by the context, we can use a number of projections appropriate in the worst case anyway.
\end{itemize}

\noindent Other key parameters used for mining operations strongly influence the system's performance, but cannot be so easily determined :
\begin{itemize}
        \item \textbf{Minimum collisions threshold.} This parameter defines the number of collisions considered as ``significant'' in terms of similarity of the corresponding subsequences. This threshold should not be too high so that we miss some tentative motifs. Given that there are next steps to refine the decision about whether a subsequence is representative of a motif, we prefer selecting ``too much'' candidates at this stage. We should however be careful that the main interest of projections is preserved : not examining the collisions between all possible subsequences all together. 
        \item \textbf{Maximum distance threshold.} This parameter set a upper bound on the actual distance between subsequences so that they are considered as similar, and eventually as representative of a same motif. A compromise needs to be found between effectively considering all subsequences representative of a same behavior as similar, even in the presence of noise (sensitivity), and not including \emph{wrong} subsequences as representative of that behavior (specificity).
\end{itemize}
According to the purpose of motifs extraction, and considering the length of representation of simulated time-series corresponding to typical activities of a person at home, we decide to consider 4 symbols to made up basic subsequences for projections. The mask used for projections is fixed of one unit length for both the number of symbols and the number of components of each symbol.

\begin{figure}[p]
\begin{center}
\includegraphics[width=14cm]{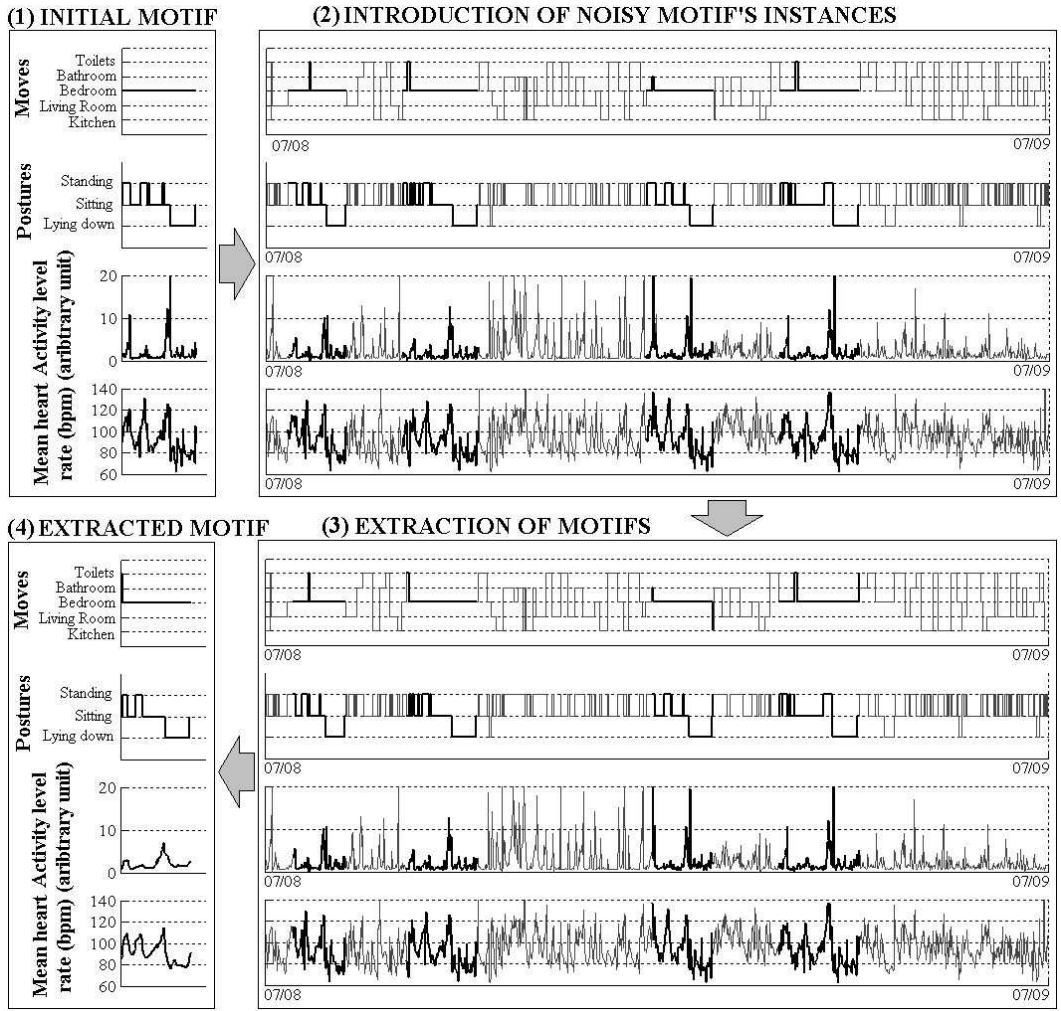}
\end{center}
\caption{\small \textbf{A sample of pattern recognition from time-series.}}\label{fig_motifs_res}
\small From left to right, and top to bottom, graphs represent (1) a reference motif, (2) introduced several times a day as noisy instances over a ``non-pattern'' sequence. Then, (3) the next graph represents frequent subsequences as extracted and classified all together from the analysis of previously defined ``pattern'' and ``non-pattern'' sequence, and at last (4) the subsequence defined as representative of motif.
\end{figure}

\subsection{Quality of the results}

\noindent First experimentations are performed using temporal sequences whose structure is known \textit{a priori}. Several instances of a motif randomly selected in simulated sequences are introduced with a moderate amount of noise in a ``non-pattern'' sequence, that is generated from random moves. Different types of noise are added to motifs' instances when introduced in non-pattern signals, that is: noisy values, interruptions, and stretching in time (see \S \ref{experimental_sets}). A sample of running this experimental process in a noisy context is illustrated through the graphes of figure \ref{fig_motifs_res}.

After checking the general performance of the system, we especially test its performances in a particularly noisy context (test of sensitivity), such as the results obtained in the presence of abnormaly modified motifs' instances (test of specificity).

\begin{figure}[p]
\begin{center}
\includegraphics[angle=90, width=14cm]{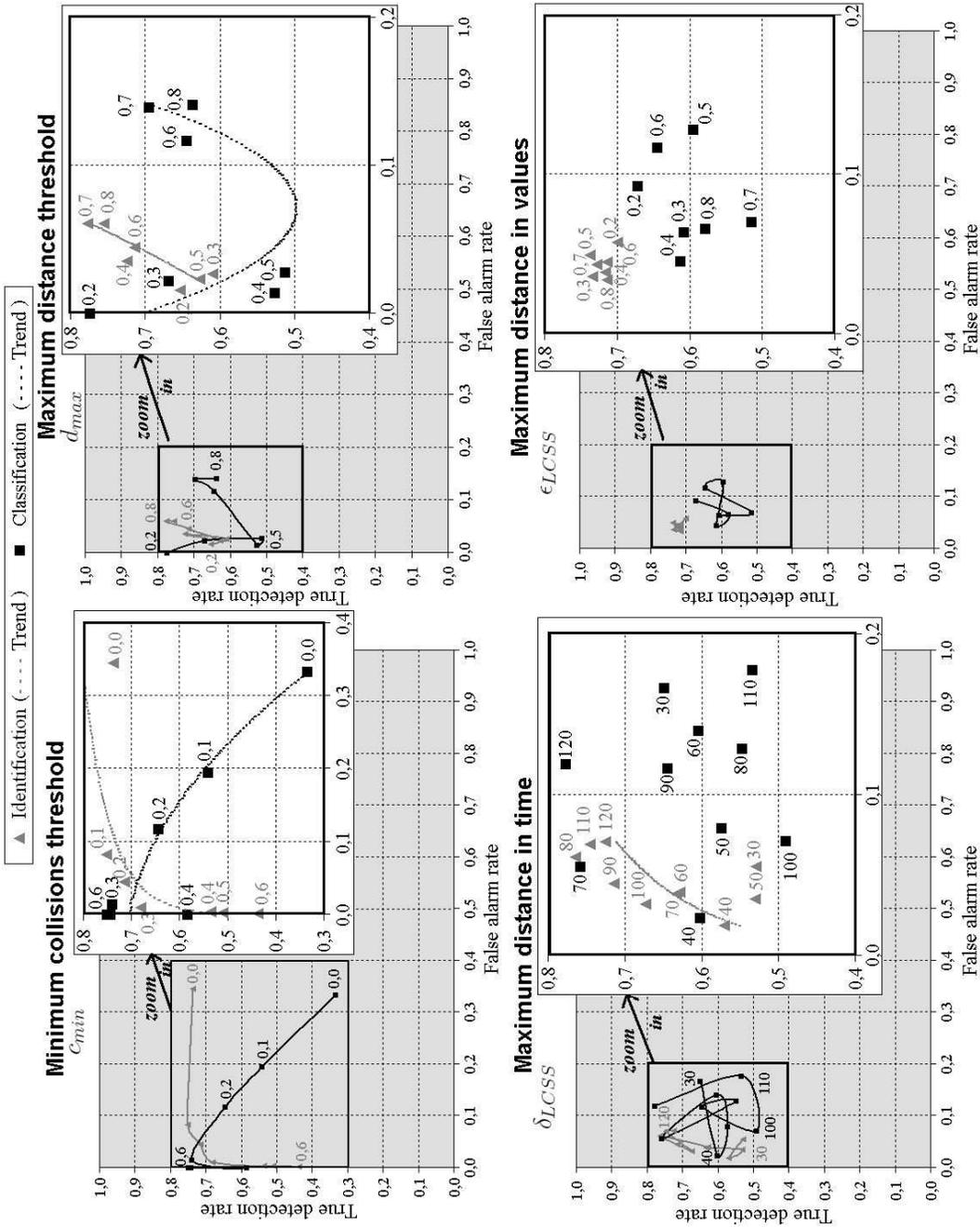}
\end{center}
\caption{\small \textbf{Mean performance results in terms of true (sensibility) and false positive (1-specificity) rates of the two steps of identifying the tentative motifs and classifying them into motifs, and considering moderate amount of noise between motifs' instances.}}\label{ROC_piti_bruit}
\small We observe individually the influence of each critical parameter in a default configuration of the others.
\end{figure}

\subsubsection{General performance results}\label{quality_results_perf}

The first experimental goal is to study the influence of the key parameters on the performances of motifs extraction, evaluated in terms of sensibility and specificity of identifying the tentative motifs and classifying them into motifs. Due to the way of selecting tentative motifs, that is from discrete and aggregated subsequences, the tentative motifs identification cannot be performed really precisely in terms of their indexes of starting and ending (sensitivity and specificity of extraction). Anyway, in our context, the purpose is to identify the occurrence of a motif's instance, without need for exact time and duration. Concerning classification, getting good performances of the system is much more fundamental. Ideally, we need for a ``perfect'' classification, even if motifs' instances are not precisely identified along time, so that we can recognize all the main activities of a person at home. The system may fail in well classifying the motifs' instances because of bad similarity measures according to the distance threshold. That may be because one of the motif's instance as been too roughly defined, including too many ``non-pattern'' points for instance and missing too many ``pattern'' ones, so that the distance increases. Then, improving the precision of tentative motifs identification may be indirectly required to get better performances of clustering. 

Critical parameters under study are the collisions and distance threshold, such as thresholds used for computing similarity measure, that is the maximum difference between two similar points in terms of values, $\epsilon_{LCSS}$, and time, $\delta_{LCSS}$. A default system configuration is then defined after many experimentations with varying values of the key parameters, while looking for the best performances in terms of classification results. The graphs of figure \ref{ROC_piti_bruit} present the system's performance according to some possible values of these thresholds. The results highlight the complexity of selecting appropriate values of these parameters, especially because of their relative influence on the system's performances. That is particularly noticed considering the results associated to varying values of \textbf{maximum distances in values} ($\epsilon_{LCSS}$) and \textbf{time} ($\delta_{LCSS}$) defining the similarity measure.

\noindent The \textbf{collisions threshold} has not a strong influence on the system's performances, and does not require to be precisely defined. In that specific case, the only interest in defining a collisions threshold lower than the number of projections is to deal with possible imprecision in the representation step. Then, if the minimum collision threshold is too high, we might miss some motifs' instances. Reversely, if too low, we possibly accept ``too much'' subsequences as potential tentative motifs, and that then requires a lower distance threshold for a proper identification of tentative motifs.

\noindent The \textbf{distance threshold} is clearly more critical since it is responsible for the end decision about whether or not a frequent subsequence is a motif. Concerning tentative motifs extraction, both true positive and false positive rates increase with the decision threshold : more motifs' instances are well identified, such as more non-pattern subsequences. On the other hand, performances related to the classification task present complex variations of true and false positive rates. Over a certain value, increasing the decision threshold gives better sensibility indexes -- more subsequences are considered as similar --  but decreases specificity ones -- in the same time, some subsequences might be considered as hazardously similar. Considering lower decision thresholds gives however better performances anyway possibly due to higher precision in identifying the motifs' instances, containing few ``non pattern'' vectors, so that similarity measures are much more relevant for classification.

\noindent Consequently, we noticed a close relation between the parameters involved in motifs extraction and classification. Then, there is some difficult compromises to be found in defining appropriate values of these parameters, especially in contexts where the system needs to support different possible noise.

Generally, according to the results presented in table \ref{tab_motifs_perfs}, the proposed approach for motifs extraction gives good results in terms of sensitivity and specificity of both extraction and classification of motifs. We however notice that the performances indexes are higher for identification than classification, as well as for specificity than sensibility. The system might sometimes fail in well identifying the whole motifs' instances, and some motifs' instances may consequently be missed in the corresponding class. Perfect classification is however possible in some cases. The large variability in the results may partially due to the random selection of motifs for each experimentation. Consequently, corresponding subsequences might not all be representative of a same level of recurrent behavior.

\begin{table}[h]
        \begin{center}
                \begin{tabular}{lccccccc}
                        \hline
                         & \multicolumn{3}{c}{\textbf{Identification}} & \multicolumn{3}{c}{\textbf{Classification}} & \textbf{Segmentation}\\
                        \emph{Indexes} & & $S_e$ & $S_p$ & & $S_e$ & $S_p$ & $\lambda$\\
                        \hline
                        \hline
                        \textbf{Mean} & & \textbf{0.71} & \textbf{0.92} & & \textbf{0.66} & \textbf{0.79} & \textbf{0.89} \\
                        Standard deviation & & 0.18 & 0.07 & & 0.34 & 0.26 & 0.19 \\
                        \hline 
                        \hline
                        \multicolumn{1}{l}{Perfect indexes}& & -- & -- & & 35\% & 60\% & 70\% \\
                        \multicolumn{4}{l}{\textbf{Perfect classification}} & \multicolumn{3}{c}{\textbf{20\%}} & \\
                        \hline
                \end{tabular}
                \end{center}
\caption{\small \textbf{Mean performance results of motifs extraction in a default system configuration and considering moderate amount of noise between motifs' instances.}}\label{tab_motifs_perfs}
\small The table presents the mean performances in terms of sensibility ($S_e$) and specificity ($S_p$) of the two steps of identifying the tentative motifs and classifying them into motifs, as well as the index of segmentation of motifs recognition ($\lambda$). All indexes fall into [0,1], the maximum values corresponding to perfect results.
\end{table}

\begin{figure}[p]
\begin{center}
\includegraphics[width=\textwidth]{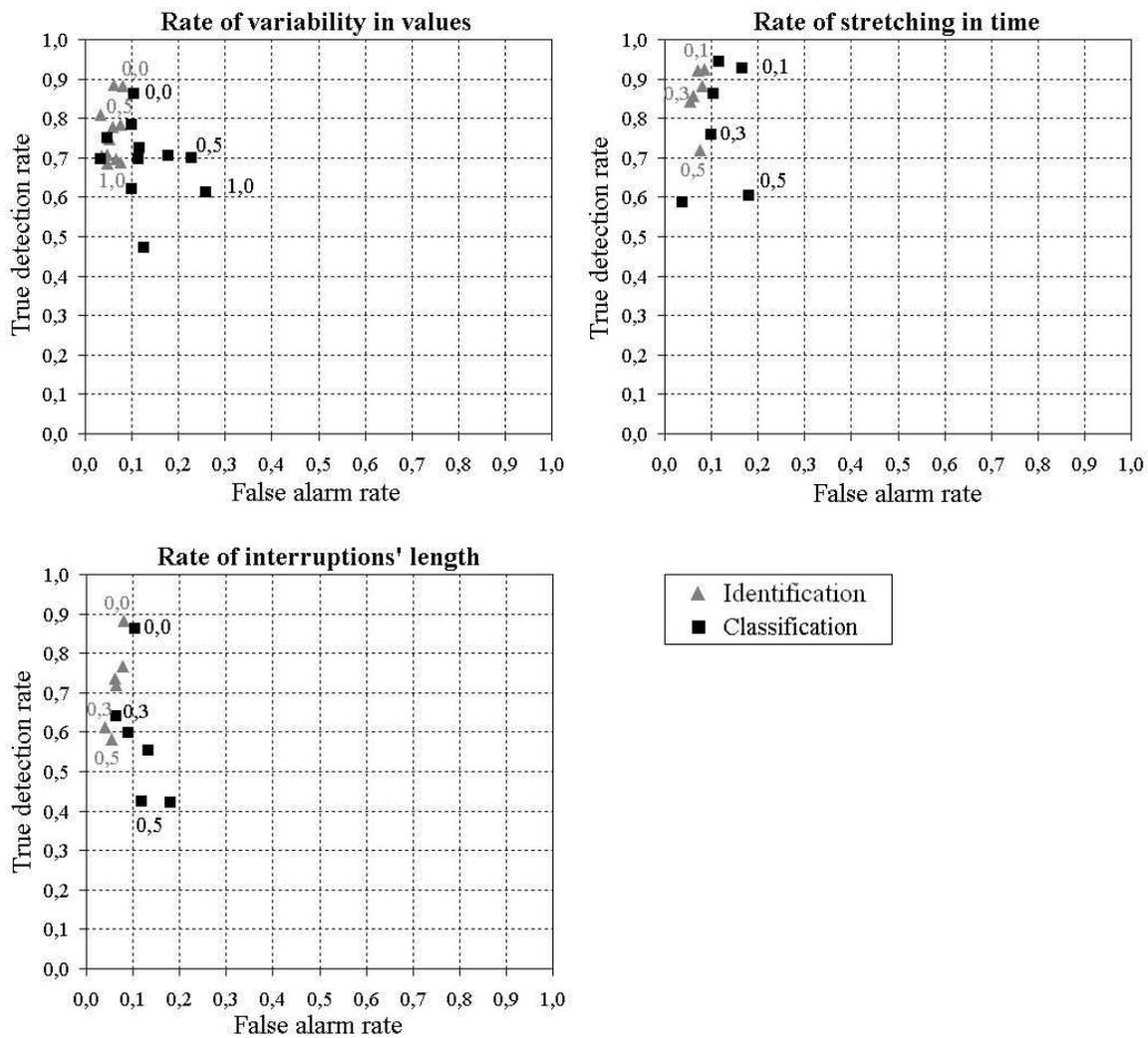}
\end{center}
\caption{\small \textbf{Mean performance results in terms of true (sensibility) and false positive (1-specificity) rates of the two steps of identifying the tentative motifs and classifying them into motifs, and considering varying rates of noise between motifs' instances.}}\label{ROC_sensibilite}
\small We observe individually the influence of each type of noise (variability in values, stretching in time, interruptions) on the performance measures in a default configuration of the system.
\end{figure}

\subsubsection{Sensitivity test}\label{quality_results_se}

The previous study gives an idea of appropriate values for the parameters of motifs extraction, so that we can observe how adding noise influences the system's performances with default values set for extraction parameters. In our experimental context, we should support certain amounts of variability in values, stretching in time, and interruptions.

Analyzing some experimental results, reported on figure \ref{ROC_sensibilite}, show good results for pattern extraction and classification even in the presence of noise. Increasing amounts of the different types of noise effectively degrades sensitiviy indexes. The system as defined using default values of extraction parameters appears to remain especially efficient in the presence of noisy values or stretching in duration, and less resistant to the presence of long interruptions in motif's instances. Anyway, in our experimental context, we do not really know at that stage where is the boundary between ``normal'' and ``abnormal'' behaviors.

These results also highlights the complexity of defining appropriate parameters of extraction to deal with all possible types of noise. On one hand, introducing large amount of noise in values or duration mainly requires to increase the values of  $\epsilon_{LCSS}$ and $\delta_{LCSS}$ constraining the similarity between points. High rates of noise in values especially need for higher $\epsilon_{LCSS}$, whereas large possible variations in duration require increasing $\delta_{LCSS}$. On the other hand, introducing interruptions in motifs' instances mainly influence the selection of appropriate collisions and distance thresholds. That indeed requires to reduce the collisions rate while increasing the distance threshold. 

\begin{figure}[p]
\begin{center}
\includegraphics[angle = 90, width=9.5cm]{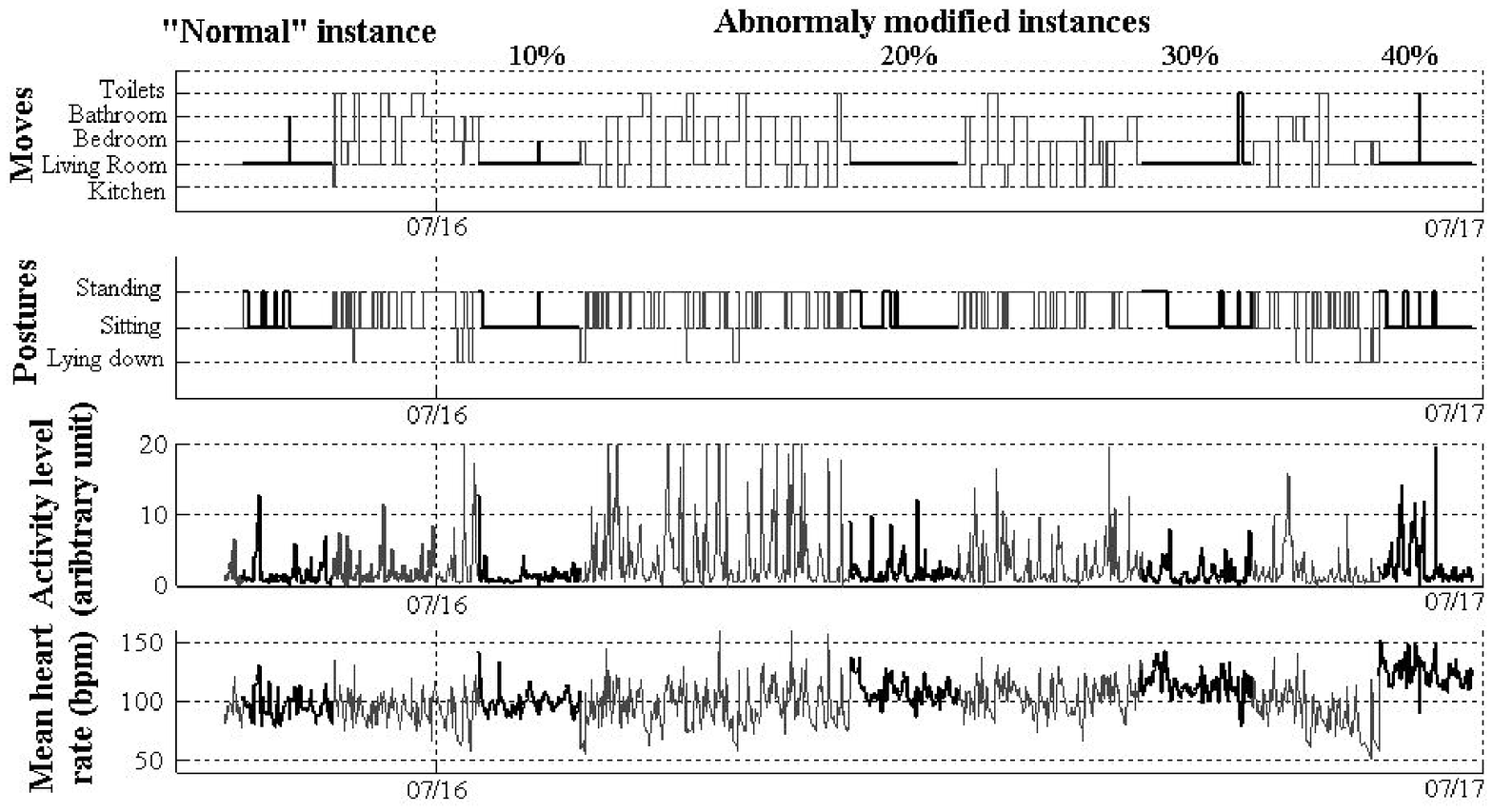}
\end{center}
\caption{\small \textbf{Sample of introducing worrying changes between motifs' instances.}}\label{motifs_dev}
\small The motifs' instances are drawn in bold type within a ``non-pattern'' sequence, including increasing rates of abnormal changes. In that sample, values of the mean heart rate are globally increased, independently of the activity, by 10 to 40\%. 
\end{figure}

\subsubsection{Specificity test}\label{quality_results_sp}

Another test required in the context of critical situation detection is to check the specificity of the system, that is the non-classification of abnormaly modified instances in the class corresponding to ``normal'' ones. Figure \ref{motifs_dev} shows a sample of introducing worrying changes in the mean heart rate features, getting to higher values independently of the activity level. Table \ref{tab_motifs_spec} reports the classification results obtained when introducing successively within a ``non-pattern'' sequence eight ``normal'' noisy instances of a given motif followed by four abnormaly modified ones. We especially notice that abnormal instances are rarely associated to the ``normal'' class. Their recognition rate as ``normal'' also decreases with increasing worrying change rate in behavior. These results validate that the system might detect critical situations.

\begin{table}[h]
\begin{center}
\begin{tabular}{lcccccc}
\hline
\textbf{Change in behavior} & \textbf{``Normal''} & & \multicolumn{4}{c}{\textbf{Worrying}}\\
\emph{Worrying change rate} & \emph{0\%} & & \emph{10\%} & \emph{20\%} & \emph{30\%} & \emph{40\%} \\
\hline\hline
\textbf{Recognition rate} & \textbf{77.5\%} & & \textbf{25\%} & \textbf{10\%} & \textbf{10\%} & \textbf{5\%} \\
\hline
\end{tabular}
\end{center}
\caption{\small \textbf{Results of classifying tentative motifs from a learning sequence containing both ``normal'' and abnormaly modified motifs' instances.}}\label{tab_motifs_spec}
\small The above table presents the mean percentage of several types of motifs' instances -- ``normal'' and abnormaly modified ones -- properly identified and classified in a so called ``normal'' class.
\end{table}

\begin{figure}[p]
\begin{center}
\includegraphics[width=\textwidth]{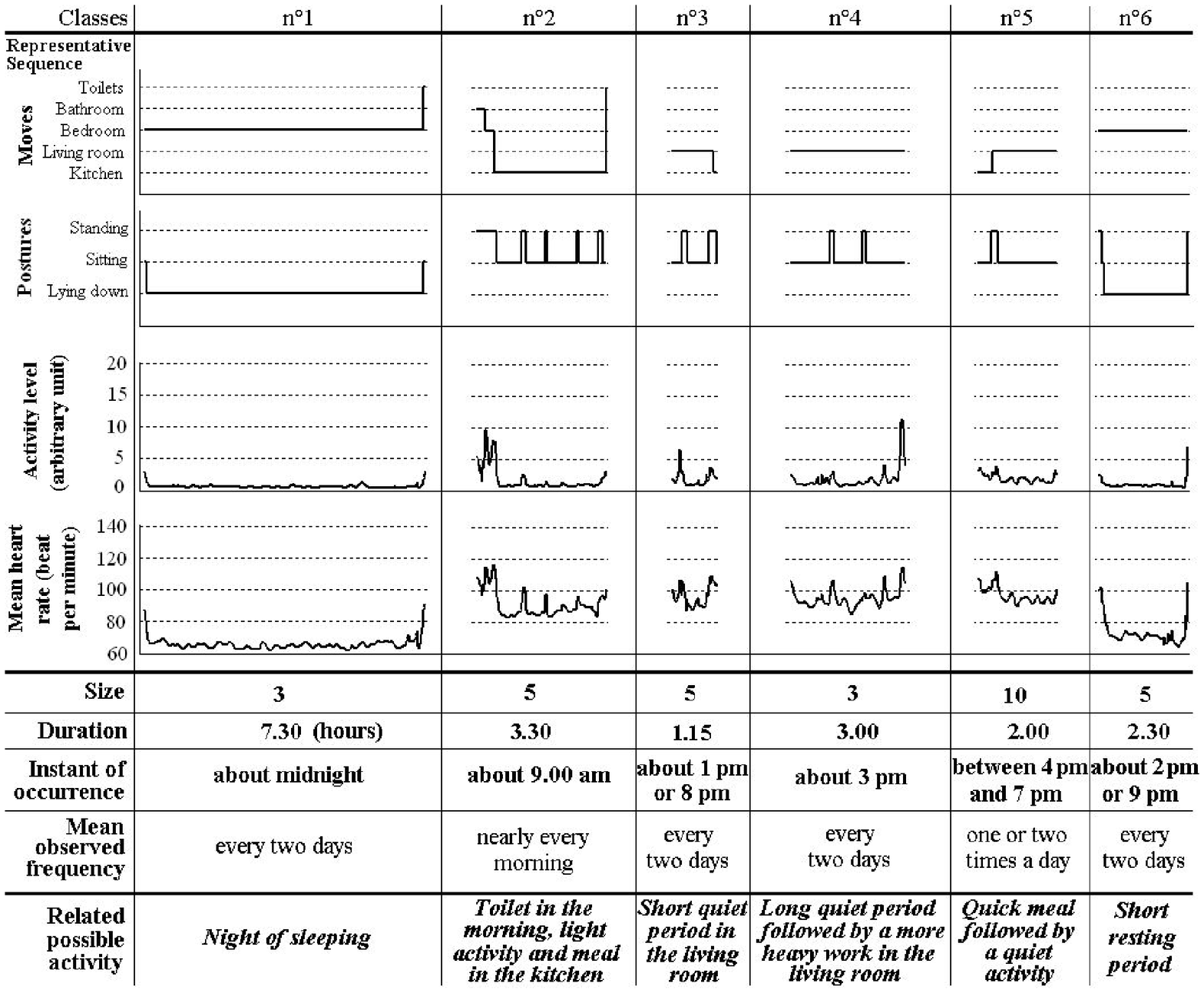}
\end{center}
\caption{\small \textbf{Features of recurrent behaviors identified within a learning sequence simulated for a given person over seven successive days.}}\label{motifs_avq}
\small A plausible interpretation of each motif is given in terms of possible activities of daily living.
\end{figure}

\subsection{Identifying reccurent behaviors from simulated sequences}

\noindent As a step of validation, we propose to observe the results of extracting motifs from a learning sequence generated over seven days by our simulation process. Simulated sequences might contain subsequences corresponding to recurrent behaviors of the person at home, even if we do not have any a priori knowledge about their explicit features.
The results reported on figure \ref{motifs_avq} highlight that the extracted motifs can be easily interpreted in terms of possible activities of daily living. The frequencies observed are however lower than expected ones, possibly due to some imprecision or incorrectness in the simulation process or to a wrong adjustment of the learning parameters. That however validates the potential good results of the proposed approach for motifs extraction in that experimental context.

\section{Conclusion and perspectives}\label{conclusion}

In that paper, we have proposed an approach for mining heterogeneous multivariate time-series to identify meaningful patterns. Generally, some interesting features of the proposed method are the ability to extract motifs from time-series containing both pattern and non-pattern subsequences, in a completely unsupervised way, allowing for noisy values between motifs' instances, and without the need for large amount of learning data sets -- the presence of two instances of a frequent pattern is theoretically enough to identify the corresponding motif.
Furthermore, the proposed approach presents several advantages compared with several works in finding time-series motifs.

\noindent First, we have considered the general case of multivariate and heterogeneous time-series. That particularly implies to define (1) an homogeneous representation of time-series to first roughly identify frequent subsequences, and (2) a similarity measure appropriate to heterogeneous multivariate time-series, in order to compare precisely these frequent subsequences. We have then extended the non-metric distance based on \emph{LCSS} experimented by Vlachos \textit{et al.} \cite{vlachos} to that general context.

\noindent Second, we have extended the \emph{projection} algorithm already experimented by Chiu \textit{et al.} \cite{chiu} for finding time-series motifs to the case of multidimensional symbols defining discrete time-series. This algorithm is indeed particularly suited to strong presence of noise. Our context of using this algorithm also differs from \cite{chiu} in defining symbols of possibly different lengths, allowing for stretching in time between discovered motifs' instances. We have also proposed a method for extending basic recurrent subsequences to the identification of representative subsequences in terms of observing daily living habits. At the end, a divisive approach to clustering allows to synthesize this set of recurrent subsequences in non-overlapping tentative motifs, then classified into motifs.

Another specificity of our approach lies in a large scale involved from the level of details embedded in raw data to the decision level. That requires to define several levels of extracting relevant information from the original time-series, up to the decision level. However, the decision's purpose considered in our experimental context does not restrict at all the possible use of our approach at different temporal scales or levels of details, or for other applications. First, the same approach can be used a different temporal scale by simply changing the frequency of raw data. Second, appropriate values of parameters can be determined to deal with other context and purpose of decision. The levels of both representation and mining operations can be adapted to a given experimental context. At the representation stage, increasing the filter length, the minimum distance threshold for allowing aggregation, and/ or reducing the number of discrete intervals entails the extraction of ``higher level'' motifs, and reversely. At the stage of mining operations, changing the values of parameters also results in modifying the sharpness of study. For instance we extract more precise frequent patterns by reducing the maximum distance threshold, or longer motifs are identified by increasing the number of symbols defining \emph{basic subsequences} for projections.
Generally, the proposed approach could be appropriate to deal with any application that aims at profiling an usual behavior from the observation of any set of complementary parameters.

First experimentations of the proposed approach for mining heterogeneous multivariate time-series give really promising results. The method allows to well identified a large number of motifs' instances introduced within non-pattern sequences, even in the presence of noise, allowing variability in values, stretching in time, and interruptions. 
Results also highlight the difficulty of selecting appropriate parameters to a given experimental context. Each possible type of noise influence mainly one type of parameters, which are however closely related one to each other. 

Additional experiments may require real data sets to first better define ``normality'' and ``abnormality'' in terms of the related temporal features, and then especially to characterize subsequences representative of recurrent behaviors, so that testing the proposed approach may be most relevant.
In the context of home health telecare, another step of experimentation is then to validate it is effectively possible to identify similar behaviors in that way from time-series recorded in realistic environments of monitoring a person at home.


\end{document}